\documentclass[journal]{IEEEtran}
\usepackage{amsmath,amsfonts}
\usepackage{algorithm2e}
\usepackage{array}
\usepackage{subcaption} 
\usepackage[caption=false,font=normalsize,labelfont=sf,textfont=sf]{subfig}
\usepackage{textcomp}
\usepackage{stfloats}
\usepackage{url}
\usepackage{verbatim}
\usepackage{graphicx}
\usepackage{cite}
\usepackage{placeins}
\usepackage{color}
\RestyleAlgo{ruled}

\usepackage{amssymb, flushend}
\newtheorem{remark}{Remark}
\begin{document}

\title{Filtering Jump Markov Systems with Partially Known Dynamics:\\ A Model-Based Deep Learning Approach}

\author{George Stamatelis,~\IEEEmembership{Student Member,~IEEE,} and George C. Alexandropoulos,~\IEEEmembership{Senior Member,~IEEE}
\thanks{This work has been supported by the SNS JU project 6G-DISAC under the EU’s Horizon Europe research and innovation program under Grant Agreement No 101139130. The contribution of G. Stamatelis has been supported by the Hellenic Foundation for Research and Innovation (HFRI) under the 5th Call for HFRI PhD Fellowships (Fellowship Number: 21080). For the experimental results, AWS resources were used which were provided by the National Infrastructures for Research and Technology (GRNET), Greece and funded by the EU Recovery and Resiliency Facility.
}
    \thanks{The authors are with the Department of Informatics and Telecommunications, National and Kapodistrian University of Athens, Panepistimiopolis Ilissia, 16122 Athens, Greece (e-mails: \{georgestamat,  alexandg\}@di.uoa.gr).} 
}



\maketitle

\begin{abstract}
This paper presents the Jump Markov Filtering
Network (JMFNet), a novel model-based deep learning framework for real-time state-state estimation in jump Markov systems with unknown noise statistics and mode transition dynamics. A hybrid architecture comprising two Recurrent Neural Networks (RNNs) is proposed: one for mode prediction and another for filtering that is based on a mode-augmented version of the recently presented KalmanNet architecture. The proposed RNNs are trained jointly using an alternating least squares strategy that enables mutual adaptation without supervision of the latent modes. Extensive numerical experiments on linear and nonlinear systems, including target tracking, pendulum angle tracking, Lorenz attractor dynamics, and a real-life dataset demonstrate that the proposed JMFNet framework outperforms classical model-based filters (e.g., interacting multiple models and particle filters) as well as model-free deep learning baselines, particularly in non-stationary and high-noise regimes. It is also showcased that JMFNet achieves a small yet meaningful improvement over the KalmanNet framework, which becomes much more pronounced in complicated systems or long trajectories. Finally, the method's performance is empirically validated to be consistent and reliable, exhibiting low sensitivity to initial conditions, hyperparameter selection, as well as to incorrect model knowledge.
\end{abstract}


\section{Introduction}
The Kalman Filter (KF) \cite{KFOriginal}, along with its extensions including the Extended KF (EKF) \cite{EKF-ref1}, are among the most well known and widely used algorithms in the signal processing community, having an extensive range of applications. In fact, despite being developed over $50$ years ago, KFs remain fundamental tools for engineering practitioners \cite{KFApplications}. In addition to traditional signal processing and control applications, these filters have recently been applied to emerging Artificial Intelligence (AI) research directions, including distributed multi-agent systems \cite{KFMultiAGentSystems}, and deep actor critic algorithms \cite{KF_actorCritic}.

The KF family constitutes Model-Based (MB) algorithms designed to estimate the hidden state of a dynamically evolving system in the presence of Gaussian noise. The term MB indicates that these algorithms require precise knowledge of the system's state transition model and the statistical properties of the process as well as that of the observation noise. However, despite their strong theoretical foundations and decades of research, traditional MB filtering approaches face several practical challenges, among which belong the following:
\begin{itemize}
\item[A)] In real-world settings, the assumed models for state evolution and observation generation are often only rough approximations of the true underlying dynamics.
\item[B)] In high-dimensional systems, the exact structure of the noise is typically unknown, non-stationary, and difficult to model accurately.
\item[C)] In nonlinear systems, methods like the EKF can become computationally expensive, particularly when dealing with high-dimensional observations.
\end{itemize}
In response to these limitations, a recently growing body of work explores KFs augmented with Neural Networks (NNs). As summarized in the recent survey~\cite{AI_KF-survey}, such approaches generally fall into the following three broad categories:
\begin{enumerate}
\item Feature-augmented filtering: Integration of NNs as feature extractors to preprocess high-dimensional observations before applying KF/EKF (e.g., \cite{KFwNeuralExtraction1}).
\item NN-enhanced filtering: Embedding of NNs directly into the filtering architecture to replace or complement classical update steps (e.g., \cite{KalmanNet, KalmanNetRobotSurgery, GraphKalmanNet}).
\item End-to-end learned dynamics: Deployment of NNs to learn complex state transition and observation models from data (e.g., \cite{deepKalmanFilters, NeuralKF_SS1, NeuralKF_SS2,foundationalMarkovJumpProc}).
\end{enumerate}

A major limitation of prior work in MB Deep Learning (DL) for filtering in Dynamical Systems (DSs) is that it typically assumes a fixed, stationary dynamical system, i.e., a single underlying transition model. This assumption restricts the applicability of such methods to systems with more complex switching behaviors or nonstationary dynamics. Jump Markov Systems (JMSs), also known as Jump Markov Linear/Nonlinear Systems, are a class of dynamical models in which the system's behavior evolves according to both continuous-valued states and a discrete-valued mode that can change over time. At a high level, JMSs capture scenarios where a system transitions between multiple distinct modes, each governed by its own set of dynamics, according to a probabilistic (typically Markovian) rule. As such, these models provide a principled framework for representing real-world systems that undergo abrupt structural changes or mode switches, such as faults, regime shifts, or operational reconfigurations. Applications of JMSs are widespread and include: target tracking (where objects switch between motion patterns)~\cite{IMMRef1,IMMRef2}; financial modeling (regime changes in markets)~\cite{RegimeSwitch}; fault detection in control systems~\cite{FaultDetectionBook}; and robotics~\cite{IMMRobotics1}. Their ability to unify discrete and continuous dynamics under uncertainty makes JMSs highly valuable for robust decision making as well as estimation tasks in complex environments. 

The most widely used class of filtering methods for JMSs is the Interacting Multiple Model (IMM) framework~\cite{IMMRef1, IMMRef2}. The core idea behind the MB IMM approach is to maintain a probability distribution over the possible modes and simultaneously run a bank of filters, typically KFs or EKFs, with each filter tailored to a specific mode. The individual estimates from these filters are then combined into a single, unified state estimate by weighting them according to the current mode probabilities. Beyond the basic IMM framework, more advanced MB estimators have been developed for linear systems where mode transitions depend on external (exogenous) inputs, as in~\cite{AISTAT-RecrurrentSwitchSystems}. However, like traditional MB KFs, IMM and its extensions rely on accurate knowledge of both the noise covariance structure and the mode transition probabilities. The assumption of fixed and perfectly known mode transition dynamics can be a significant limitation in real-world applications.

In this paper, inspired by the recently proposed NN-enhanced filtering approaches and specifically the KalmanNet architecture \cite{KalmanNet}, we focus on the problem of state estimation in JMSs under partial domain knowledge, leveraging the representational power of NNs. In particular, we assume that the state transition and observation models for each mode are either exactly known or approximately specified, while the noise covariance structures are unknown. Additionally, we do not assume any knowledge of the mode transition probabilities, which may be unknown and nonstationary. To tackle this challenging setting, we propose the Jump Markov Filtering Network (JMFNet) which is a hybrid NN architecture composed of two key NN components:
\begin{itemize}
\item[\textit{i})] Mode Prediction NN: A deep NN model that maps a history of observations to a probability distribution over the discrete modes.
\item[\textit{ii})] Mode-Augmented KF: An NN that extends the KalmanNet framework by incorporating an one-hot encoding of the predicted mode as an additional input.
\end{itemize}
The two NNs are jointly trained using an Alternating Least Squares (ALS) optimization strategy, enabling mutual adaptation of mode inference and state estimation. Through extensive numerical evaluation, it is reveled that our method can outperform both purely MB filters, Model-Free (MF) NNs as well as the transition-agnostic KalmanNet architecture. The performance of the proposed is assessed over a wide range of applications, including linear and nonlinear target tracking, harmonic pendulum movement, chaotic systems, and real-world datasets.

\textit{\textbf{Notation:}} Throughout this paper, lowercase letters refer to scalar parameters, e.g., $x$, and bold letters refer to vectors, e.g., $\mathbf{x}$. Bold uppercase letters, e.g., $\mathbf{X}$ are reserved for matrices, and calligraphic letters, e.g., $\mathcal{X}$, represent sets. The letter $t$ is typically reserved for discrete time instances, and also define $\mathbf{x}_{1:t}\triangleq\{\mathbf{x}_1,\mathbf{x}_2,\ldots,\mathbf{x}_t\}$, and $||\mathbf{x}||_2$ denotes the Euclidean norm of vector $\mathbf{x}$. Finally, $\mathcal{N}(\mathbf{m},\mathbf{\Sigma})$ is a multivariate normal distribution with mean $\mathbf{m}$ and covariance matrix $\mathbf{\Sigma}$.

\section{Background on State-Space Models}
In the section, we overview single- and multi-mode State-Space (SS) models and their most representative estimation algorithms.

\subsection{Single-Mode Dynamical Systems (DSs)}
\subsubsection{Linear DSs (LDSs)}
Such systems are typically modeled as discrete-time SS
models\cite{SSBook}, with the simplest of them being the LDSs, where a state vector $\mathbf{x}_t \in \mathbb{R}^{s\times1}$ evolves according to a linear update rule, and an observation vector $\mathbf{y}_t \in \mathbb{R}^{o\times1}$ is a noisy linear function of $\mathbf{x}_t$. More specifically\footnote{In this paper, we focus on the control-free LDSs, however, the proposed methods can be trivially extended to the case where the state update incorporates an additive control term $\mathbf{B} a_t$, where $a_t$ is a known control signal.}: 
\begin{subequations}\label{Lds}
\begin{align}
    \mathbf{x}_t &= \mathbf{F} \mathbf{x}_{t-1}+\mathbf{w}_t, & \mathbf{w}_t &\sim \mathcal{N}(\mathbf{0},\mathbf{Q}), \label{Lds-x} \\
    \mathbf{y}_t &= \mathbf{H} \mathbf{x}_t + \mathbf{v}_t, & \mathbf{v}_t &\sim \mathcal{N}(\mathbf{0},\mathbf{R}), \label{Lds-y}
\end{align}
\end{subequations}
where $\mathbf{F,Q} \in \mathbb{R}^{s \times s}$, $\mathbf{H} \in \mathbb{R}^{o \times s}$, and $\mathbf{R} \in \mathbb{R}^{o\times o}$. One common application of this model is target tracking, where $\mathbf{x}_t$ represents a target's position, velocity, and acceleration. When several sensors are placed in various areas in the target's vicinity, they can collect information encapsulated in the vector $\mathbf{y}_t$. The noise vectors $\mathbf{w}_t$ and $\mathbf{v}_t$ model respectively uncertainty in the state evolution model and the sensor reliability.

The focus in this paper is on filtering, i.e., the online estimation of the hidden state $\mathbf{x}_t$ given the past and current observations $\mathbf{y}_{1:t}$. Other tasks, such as smoothing (i.e., offline estimation of entire state sequences), missing observation recovery, and future observation prediction, are left for future investigation.

\subsubsection{Kalman Filter (KF)}
This filter is a MB linear estimator that maps the most recent a posteriori estimate $\hat{\mathbf{x}}_{t-1|t-1}$ and the current observation $\mathbf{y}_t$  to an estimate of the true state $\mathbf{x}_t$. To achieve this goal, KF requires knowledge of both the state update functions $\mathbf{F}$ and $\mathbf{H}$, as well as the covariance matrices $\mathbf{Q}$ and $\mathbf{R}$. It has been shown that KF is the optimal linear estimator for the Mean Squared Error (MSE) minimization criterion when $\mathbf{Q}$ and $\mathbf{R}$ are known~\cite{KalmanOptim}. Specifically, at each $t$, KF performs the following two operations:

\textbf{1. Prediction step:} Firstly, the current state and observation are predicted, along with uncertainty metrics for these estimates: $\mathbf{\Sigma}_{t|t-1} \in \mathbb{R}^{s \times s}$ and $\mathbf{S}_{t|t-1} \in \mathbb{R}^{o \times o}$ respectively. The current a priori estimates are formed by using only the most recent a posteriori estimates, achieving constant computational complexity over time. The update rules are given as follows:
\begin{subequations}\label{KF-pred}  
\begin{align}
    \hat{\mathbf{x}}_{t|t-1} &\gets \mathbf{F} \hat{\mathbf{x}}_{t-1|t-1}, \label{KF-pred-x} \\
    \mathbf{\Sigma}_{t|t-1} &\gets \mathbf{F} \mathbf{\Sigma}_{t-1|t-1} \mathbf{F}^T + \mathbf{Q}, \label{KF-pred-Sigma} \\
    \hat{\mathbf{y}}_{t|t-1} &\gets \mathbf{H} \hat{\mathbf{x}}_{t|t-1}, \label{KF-pred-y} \\
    \mathbf{S}_{t|t-1} &\gets \mathbf{H} \mathbf{\Sigma}_{t|t-1} \mathbf{H}^T + \mathbf{R}. \label{KF-pred-S}
\end{align}
\end{subequations}

\textbf{2. Update step:} At this step, the most recent observation vector $\mathbf{y}_t$ is processed and used to compute the a posteriori estimates, as follows:
\begin{subequations}\label{eq:KalmanPosterior}
\begin{align}
        &\hat{\mathbf{x}}_{t|t} \gets \hat{\mathbf{x}}_{t|t-1} + \mathbf{K}_t  \Delta \mathbf{y}_t, \label{KF-post-x}& \\
        &\mathbf{\Sigma}_{t|t} \gets \mathbf{\Sigma}_{t|t-1} - \mathbf{K}_t \mathbf{S}_{t|t-1} \mathbf{K}_t^T \label{KF-postSigma} ,
    \end{align}
    \end{subequations}
where the \textit{Kalman Gain} (KG) $\mathbf{K}_t $ is given by:
\begin{equation}\label{eq:KalmanGain}
    \mathbf{K}_t \triangleq \mathbf{\Sigma}_{t|t-1} \mathbf{H}^T \mathbf{S}^{-1}_{t|t-1}\in \mathbb{R}^{s \times o},
\end{equation}
and the \textit{innovation} $\Delta \mathbf{y}_t$ as: 
\begin{equation}\label{eq:Inovation}
    \Delta \mathbf{y}_t \triangleq \mathbf{y}_t - \hat{\mathbf{y}}_{t|t-1}.
\end{equation}

\subsubsection{Nonlinear Models and Extended KF (EKF)}
Although LDSs are thoroughly studied in literature and they have multiple practical applications, the dynamics of the real-world applications are often not linear. In more general cases, the SS can be expressed as:
\begin{subequations}\label{eq:NonLDS}
 \begin{align}
    \mathbf{x}_t &= \mathbf{f}\left(\mathbf{x}_{t-1}\right)+\mathbf{w}_t,\label{eq:NonLDS_1}\\ 
    \mathbf{y}_t &= \mathbf{h}\left(\mathbf{x}_t\right)+\mathbf{v}_t,\label{eq:NonLDS_2}
\end{align}
\end{subequations}
where $\mathbf{f}(\cdot)$ and $\mathbf{h}(\cdot)$ are nonlinear functions. For such models, the MB KF can be replaced by the EKF updates:
\begin{subequations}\label{eqLEKF-updates}
\begin{align}
    \hat{\mathbf{x}}_{t|t-1} &\gets \mathbf{f}\left(\hat{\mathbf{x}}_{t-1|t-1}\right), \label{EKF-xupd}\\
    \hat{\mathbf{y}}_{t|t-1} &\gets \mathbf{h}\left(\hat{\mathbf{x}}_{t|t-1}\right). \label{EKF-yupd}
\end{align}
\end{subequations}
However, due to the nonlinearity of the system, the updated covariances  cannot be computed in closed form and must instead be approximated. To this end, the Jacobians:
\begin{equation}\label{eq:Jacobians}
    \hat{\mathbf{F}}_t = \mathcal{J}_{\mathbf{f}}\left(\hat{\mathbf{x}}_{t-1|t-1}\right), \quad \hat{\mathbf{H}}_t = \mathcal{J}_{\mathbf{h}}\left(\hat{\mathbf{x}}_{t|t-1}\right), 
\end{equation}
are evaluated at the current state estimates and used in the update rules (\ref{KF-pred-Sigma}) (\ref{KF-pred-S}),  (\ref{KF-postSigma}), and (\ref{eq:KalmanGain}) in place of $\mathbf{F}$ and $\mathbf{H}$. Unlike its linear counterpart, EKF is a sub-optimal filter. Besides, EKF, other approximate filtering algorithms, like the Unscented Kalman filter (UFK), have been proposed \cite{UKF}.

\subsubsection{The KalmanNet Framework}
It constitutes an interpretable, data-driven framework for performing filtering in linear and nonlinear recursive models with partial domain knowledge~\cite{KalmanNet,KalmanNetRobotSurgery}. KalmanNet leverages the structure of the system described in~(\ref{eq:NonLDS_1}) and~\eqref{eq:NonLDS_2}, and the EKF update rules in~(\ref{eqLEKF-updates}), but unlike purely MB approaches, it does not require information about the covariance matrices. At each time step $t$, the a priori estimates of $\mathbf{x}_t$ and $\mathbf{y}_t$ are computed following the EKF method. A feature vector $\mathbf{I}_t$ is then extracted and passed to a Gated Recurrent Unit (GRU) architecture $\boldsymbol\theta$\cite{GRUPaper}, which produces an estimate of the KG, $\tilde{\mathbf{K}}_t^{\boldsymbol\theta}$. The learned gain is then used to form posterior estimates in place of the actual $\mathbf{K}_t$, i.e.:
\begin{equation}
    \label{eq:posteriorKalmanNet}
    \hat{\mathbf{x}}_{t|t} \gets \hat{\mathbf{x}}_{t|t-1}+\mathbf{\tilde{K}}_t^{\boldsymbol\theta} \Delta \mathbf{y}_t.
\end{equation}
It is noted that KalmanNet does not maintain or update covariance matrices explicitly, a feature that makes it applicable to environments with complicated and non-stationary noise.

\subsection{Multi-Mode Dynamical Systems (DSs)}\label{subsec:jump_system_model}
\subsubsection{Jump Markov Linear Systems}
Let a system evolve under $M$ possible LDS models (or modes) according to the following SS model~\cite{JMLSBook}:
\begin{subequations}\label{eq:JMLS-system}
\begin{align}
    \mathbf{x}_t &= \mathbf{F}^{(j_t)} \mathbf{x}_{t-1} + \mathbf{w}_t^{(j_t)}, & \mathbf{w}_t^{(j_t)} \sim \mathcal{N}(\mathbf{0}, \mathbf{Q}^{(j_t)}), \label{eq:x_model-jump}\\
    \mathbf{y}_t &= \mathbf{H}^{(j_t)} \mathbf{x}_t + \mathbf{v}_t^{(j_t)}, & \mathbf{v}_t^{(j_t)} \sim \mathcal{N}(\mathbf{0}, \mathbf{R}^{(j_t)}), \label{eq:y_model-jump}
\end{align}
\end{subequations}
where the mode $j_t$ is a finite Markov Chain (MC), with a transition probability matrix $\bf\Pi$, i.e., $\mathbf{\Pi}_{i,j}\triangleq {\rm Pr}[j_t=j|j_{t-1}=i]$. To simplify notation, we will henceforth remove the time subscript $t$ from the mode-dependent quantities when the time instance is clear from te context. For instance, we will write $\mathbf{x}_t = \mathbf{F}^{(j)}\mathbf{x}_{t-1}+\mathbf{w}_t^{(j)}$ instead.

\subsubsection{The Interacting Multiple Model (IMM) filter}\label{subsec:IMM}
This algorithm maintains a mode \textit{belief vector} $\boldsymbol{\mu} \in \mathbb{R}^M$ representing the probability of each discrete system mode~\cite{IMMRef1,IMMRef2}. At each time instance $t$, IMM first performs a mixing step according to which initial conditions for each mode-specific KF are computed by probabilistically combining the previous state estimates using the mode transition matrix. Then, IMM runs $M$ independent KFs (one per mode) using these mixed initializations. After observing $\mathbf{y}_t$, it updates each mode's likelihood and refines the mode probabilities accordingly. Finally, the state estimates from all filters are fused into a single estimate using the updated mode probabilities as weights. In more detail:

\textbf{1. Mixing:} Initially, the a priori probabilities of being in each discrete mode are computed, along with the corresponding weighted state and covariance. The update rules are defined as follows:
\begin{subequations}    
\begin{align}\label{eq:MixingStepIMM}
    \boldsymbol{\mu}_{t-1|t-1}^{(i|j)} & \leftarrow \frac{\mathbf{\Pi}_{ij} \boldsymbol{\mu}_{t-1|t-1}^{(i)}}{c_j}, \quad
    c_j = \sum_{i=1}^M \mathbf{\Pi}_{ij} \boldsymbol{\mu}_{t-1|t-1}^{(i)},  \\
    \hat{\mathbf{x}}_{t-1|t-1}^{(0,j)} & \leftarrow \sum_{i=1}^M \boldsymbol{\mu}_{t-1|t-1}^{(i|j)} \hat{\mathbf{x}}_{t-1|t-1}^{(i)},   \\
    \mathbf{\Sigma}_{t-1|t-1}^{(0,j)} &\leftarrow \sum_{i=1}^M \boldsymbol{\mu}_{t-1|t-1}^{(i|j)} \left[ \mathbf{\Sigma}_{t-1|t-1}^{(i)} + \right. \nonumber \\
    &\left. (\hat{\mathbf{x}}_{t-1|t-1}^{(i)} - \hat{\mathbf{x}}_{t-1|t-1}^{(0,j)})(\hat{\mathbf{x}}_{t-1|t-1}^{(i)} - \hat{\mathbf{x}}_{t-1|t-1}^{(0,j)})^T \right].
\end{align}
\end{subequations}
The weighted state vectors $\hat{\mathbf{x}}_{t-1|t-1}^{(0,j)}$ will be provided as input to the KFs in the next step.

\textbf{2. Per Mode Model-Matched Filtering:} The next step involves a loop over each mode $j$. The $j$-th KF operates on the $j$-th weighted estimate of the previous step as follows:
\begin{subequations}
    \begin{align}\label{IMM-Filtering}
    \hat{\mathbf{x}}_{t|t-1}^{(j)} &\leftarrow \mathbf{F}^{(j)} \hat{x}_{t-1|t-1}^{(0,j)},    \\
    \mathbf{\Sigma}_{t|t-1}^{(j)} &\leftarrow \mathbf{F}^{(j)} \mathbf{\Sigma}_{t-1|t-1}^{(0,j)} (\mathbf{F}^{(j)})^T + \mathbf{Q}^{(j)},  \label{eq:CovPredIMM} \\
    \hat{\mathbf{y}}_{t|t-1}^{(j)} &\leftarrow \mathbf{H}^{(j)} \hat{x}_{t|t-1}^{(j)}, \label{eq:ObsPredIMM} \\
    \mathbf{S}_{t|t-1}^{(j)} &\leftarrow \mathbf{H}^{(j)} \mathbf{\Sigma}_{t|t-1}^{(j)} (\mathbf{H}^{(j)})^T + \mathbf{R}^{(j)}, \label{eq:CovPredIMM2}  \\
    \mathbf{K}_t^{(j)} &\leftarrow \mathbf{\Sigma}_{t|t-1}^{(j)} (\mathbf{H}^{(j)})^T (\mathbf{S}_{t|t-1}^{(j)})^{-1}, \label{eq:IMM-KG}\\
    \hat{\mathbf{x}}_{t|t}^{(j)} &\leftarrow \hat{\mathbf{x}}_{t|t-1}^{(j)} + \mathbf{K}_t^{(j)} (\mathbf{y}_t - \hat{\mathbf{y}}_{t|t-1}^{(j)}), \label{eq:IMM-state_update}\\
    \mathbf{\Sigma}_{t|t}^{(j)} &\leftarrow \mathbf{\Sigma}_{t|t-1}^{(j)} - \mathbf{K}_t^{(j)} \mathbf{S}_{t|t-1}^{(j)} (\mathbf{K}_t^{(j)})^T. \label{eq:CovPosteriorIMM}
\end{align}
\end{subequations}

\textbf{3. Mode Probability Update:} In the third step, the observed (true) $\mathbf{y}_t$ is used to compute the probability of being in each mode, as follows:
\begin{subequations}    
\begin{align}\label{ModeProbUpdates}
    \Lambda_t^{(j)} & \leftarrow \mathcal{N}(\mathbf{y}_t; \hat{\mathbf{y}}_{t|t-1}^{(j)}, \mathbf{S}_{t|t-1}^{(j)}),  \\
    \boldsymbol{\mu}_{t|t}^{(j)} &\leftarrow \frac{ \Lambda_t^{(j)} c_j }{ \sum_{\ell=1}^M \Lambda_t^{(\ell)} c_\ell }.
\end{align}
\end{subequations}

\textbf{4. Fusion:} Finally, the per-mode estimates $\hat{\mathbf{x}}_{t|t}^{(j)}$ of the KFs of step 2 and the updated mode probabilities of step 3 are fused to compute the final state and estimates as follows:
\begin{subequations}
    \begin{align}\label{IMM-posterior-fusion}
    \hat{\mathbf{x}}_{t|t} &\leftarrow \sum_{j=1}^M \boldsymbol{\mu}_{t|t}^{(j)} \hat{\mathbf{x}}_{t|t}^{(j)},  \\
    \mathbf{\Sigma}_{t|t} &\leftarrow \sum_{j=1}^M \boldsymbol{\mu}_{t|t}^{(j)} \left[ \mathbf{\Sigma}_{t|t}^{(j)} + (\hat{\mathbf{x}}_{t|t}^{(j)} - \hat{\mathbf{x}}_{t|t})(\hat{\mathbf{x}}_{t|t}^{(j)} - \hat{\mathbf{x}}_{t|t})^T \right].
\end{align}
\end{subequations}
In case of nonlinear switch systems, the IMM algorithm can be combined with the EKF updates replacing the linear updates of Step 2.  In this case, the state and observation updates of expressions (13a) and (\ref{eq:ObsPredIMM}) need to be modified to: \begin{equation}\label{eq:IMM-nonLinearUpdates}
   \hat {\mathbf{x}}_{t|t-1}^{(j)} \leftarrow \mathbf{f}^{(j)}\left(\mathbf{x}_{t-1|t-1}^{(0,j)}\right) \quad \text{and} \quad \hat{\mathbf{y}}_{t|t-1}^{(j)} \leftarrow \mathbf{h}^{(j)}\left(\hat{\mathbf{x}}_{t|t-1}^{(j)}\right).
\end{equation}
The Jacobian matrices $\hat{\mathbf{F}}_t$ and $\hat{\mathbf{H}}_t$ are then plugged into (\ref{eq:CovPredIMM}), (\ref{eq:CovPredIMM2}), (\ref{eq:IMM-KG}), and (\ref{eq:CovPosteriorIMM}), while the rest of the steps remain unchanged. 
\remark{Even for the linear system case, the IMM algorithm is not optimal.}

\section{Data-Driven Filters with Partial Knowledge}
Our intention, in this paper is to design an efficient mechanism that maps the past and present observations, $\mathbf{y}_{1:t}$, to an estimate of the true current state $\mathbf{x}_t$. For instance, in a vehicle tracking application, where $\mathbf{y}_t$ is a noisy estimate of the vehicle's location, the designed filtering mechanism aims to estimate the vehicle's true location, velocity, and acceleration. Our main focus is on applications with partial domain knowledge. To this end, 
we make the following assumptions: 
\begin{enumerate}
    \item Approximations of the system dynamics (i.e., the $\mathbf{f}(\cdot)$ and $\mathbf{h}(\cdot)$ functions) are available, provided perhaps by an application expert. Alternatively, they can also be computed by using more computationaly intensive offline models. Techniques for estimating these dynamics with DL, e.g., \cite{deepKalmanFilters}, constitute a very interesting research direction outside the scope of this paper.
    \item The noise distributions are unknown.
    \item The transition matrix $\mathbf{\Pi}$ is unknown and it can be non-stationary.
\end{enumerate}

We finally assume access to a large labeled dataset of sequences of observations $\mathbf{y}_t$, and their corresponding true continuous hidden states $\mathbf{x}_t$. However, labels of the true MC modes $j_t$ are assumed unavailable.

\subsection*{Why it is hard to know $\mathbf{\Pi}$?}
The IMM algorithm requires full knowledge of the mode transition matrix $\mathbf{\Pi}$. It maintains and updates a set of mode probabilities to account for uncertainty over the system's discrete dynamics. However, in high-dimensional or highly nonlinear environments, this approach can become computationally burdensome and may not capture complex switching behavior accurately. In this paper, we deploy two NNs to estimate mode probabilities and hidden state values without the knowledge of $\mathbf{\Pi}$. In fact, we do not assume a fixed stationary transition matrix.  In many real-world systems, especially those with complex or uncertain dynamics (e.g., human behavior, varying terrain, or changing environmental conditions), the true mode transition probabilities are unknown or non-stationary. For instance, consider a vehicle tracking application, where the vehicle switches between different modes for constant speed, acceleration, and turning. In certain parts of the road, the vehicle will move mostly on a straight line, whereas in other parts it will mostly turn. Consequently, the mode transition dynamics are not fixed, and hardcoding a Markov transition matrix can severely harm estimation performance.

All in all, our goal, in this paper, is to develop a MB DL method that retains the structured interpretability of a hybrid system (with a finite set of well defined dynamic modes), while at the same time, it avoids the unrealistic assumption of a known and fixed transition matrix $\mathbf{\Pi}$.

\subsection*{Why unlabeled modes?}
Consider a vehicle tracking application where the objective is to infer its true position, velocity, and acceleration from noisy position measurements. A large dataset of ground-truth state trajectories can be collected using high-precision positioning sensors deployed across various locations in the field of interest. These sensors can provide accurate measurements of dynamic quantities, such as velocity and acceleration, enabling the reconstruction of continuous-state trajectories for training. However, the discrete mode —corresponding to high-level behaviors or maneuver types (e.g., turning left, accelerating, reversing)— is not directly observable and is typically not labeled in real-world datasets. Annotating such modes would require extensive manual effort from domain experts and is often infeasible at scale. This motivates the need for machine learning methods that can infer latent mode dynamics from observed data without relying on explicit mode supervision.

\section{Proposed MB Deep Learning Framework}
In this section, we present JMFNet, a novel interpretable MB DL methodology for real-time state estimation in nonlinear JMSs with partial domain knowledge. Our approach leverages an ALS training procedure for two collaborating RNNs, which jointly infer system states and mode probabilities. Unlike traditional methods like IMM, we do not assume a fixed or known mode transition matrix. Instead, our method adaptively learns mode dynamics from data, while preserving the known state update structure and number of modes.

\subsection{NN Structures}
Two RNNs, denoted as $\boldsymbol{\theta}_m$ and $\boldsymbol{\theta}_K$, are deployed to offer mode inference and KF-like state estimation, respectively. Inspired by the structure of the classical IMM filter~\cite{IMMRef1,IMMRef2}, the mode prediction network $\boldsymbol{\theta}_m$ estimates the probability distribution over discrete modes, while $\boldsymbol{\theta}_K$ acts as a learned, mode-specific KF surrogate. For each mode, the latter NN processes the observations and a mode indicator to produce a mode-conditioned state estimate. These mode-specific estimates are then fused using the mode probabilities to form the final state prediction. To ensure architectural consistency with prior works such as~\cite{KalmanNet,DANSE}, we adopt moderate-sized GRUs~\cite{GRUPaper} as the core recurrent component in both networks. However, other structures could be used, such as Long Short Term Memory(LSTM) networks~\cite{LSTM}, and even Transformers~\cite{attentionIsAllYouNeed}. 

\paragraph{Mode Prediction NN}
The network $\boldsymbol{\theta}_m$ maps the history of observations $\mathbf{y}_{1:t}$ to a categorical distribution over discrete modes at each time instance $t$. Specifically, it produces mode probabilities as follows:
\begin{equation}
\boldsymbol{\mu}_{t,\boldsymbol{\theta}_m}^{(j)} = {\rm Pr}[j_t = j \mid \mathbf{y}_{1:t}; \boldsymbol{\theta}_m],
\end{equation}
where $j_t$ denotes the latent mode at time $t$. The most recent observation $\mathbf{y}_t$ is first transformed via a feed-forward layer to produce a feature vector $\mathbf{f}_{t,m}$. This feature vector, together with the previous hidden state $\mathbf{h}_{t-1,m}$, is then passed through a GRU to yield a hidden representation $\mathbf{o}_{t,m}$. Finally, $\mathbf{o}_{t,m}$ is processed through a softmax-activated linear layer to produce the mode probability vector $\boldsymbol{\mu}_{t,\boldsymbol{\theta}_m}$.

\paragraph{Mode-Conditioned KG Estimation NN}
The mode-augmented architecture $\boldsymbol{\theta}_K$ is inspired by the structure of the original KalmanNet~\cite{KalmanNet}, with the key distinction that it is explicitly conditioned on the current discrete mode. Instead of performing KG estimation for a stationary SS model, the proposed NN now needs to handle $M$ different state evolution models. At each time instance $t$, it receives as input an information vector $\mathbf{I}_t^{(j)}$,  
composed of two components: the KF features $\mathbf{I}^{(j)}_{t,K}$ as proposed in~\cite{KalmanNet}, and an one-hot encoded vector\footnote{One-hot means that, for each mode $m$, the vector  
has exactly one '1' in the $m$-th dimension.} indicating the current mode. 
Conditioning on this mode vector enables the NN to approximate mode-specific gain mappings within a unified parameterization.

In our implementation, we have encoded the KF–related information using the following input vector:
\begin{equation}\label{eq:InputType1}
    \mathbf{I}^{(j)}_{t,K}\!=\!\{\mathbf{y}_t\!-\!\mathbf{y}_{t-1}, \mathbf{y}_t\!-\!\hat{\mathbf{y}}^{(j)}_{t|t-1},\hat{\mathbf{x}}_{t-1|t-1}^{(j)}\!-\!\hat{\mathbf{x}}_{t-1|t-2}^{(j)}\}.
\end{equation}
This structure effectively captures both the recent observation information, prediction error, and state update dynamics, providing the NN with a compact representation of the temporal variation in the observations and latent state. Note that the same feature configuration was also found effective for non-switching neural filters in \cite{KalmanNet}.

The remainder of the architecture is similar to the mode prediction NN; in this case, the output is a two-dimensional array approximating the KG of the current mode (see~(\ref{eq:IMM-KG})). Let $\mathbf{K}_{t,\boldsymbol{\theta}_K}^j$ denote this output for mode $j$. This is plugged in the state update in~(\ref{eq:IMM-state_update}) of the IMM algorithm.
\subsubsection{Parameter Sharing between Modes}
In the proposed design, a single KG estimator is shared across all modes, instead of training $M$ independent RNNs. This choice is motivated by the following two complementary factors: \begin{itemize}
    \item \textbf{Redundancy minimization:} Modes can exhibit similar local dynamics (e.g., same angular velocity with different straight line velocity, or same velocity/acceleration on one axis in target tracking applications). Henceforth, using a shared model prevents redundant learning of similar functions. Thus, the one-hot input enables the NN to learn both shared and mode-specific behaviors.
    \item \textbf{Decreased computational and memory footprint: } Sharing parameters decreases the overall number of learnable weights from $O(M |\boldsymbol{\theta}_K|)$ to $O(|\boldsymbol{\theta}_K|)$, where $|\boldsymbol{\theta}_K|$ denotes the number of NN weights, resulting in lowering both storage requirements (which can be crucial when filtering is performed on lightweight devices (e.g., Internet-of-Things (IoT)) and the effective search space during optimization.
\end{itemize}
\paragraph{Overall Filtering Operation}
\begin{figure*}
    \scalebox{0.55}{\includegraphics[]{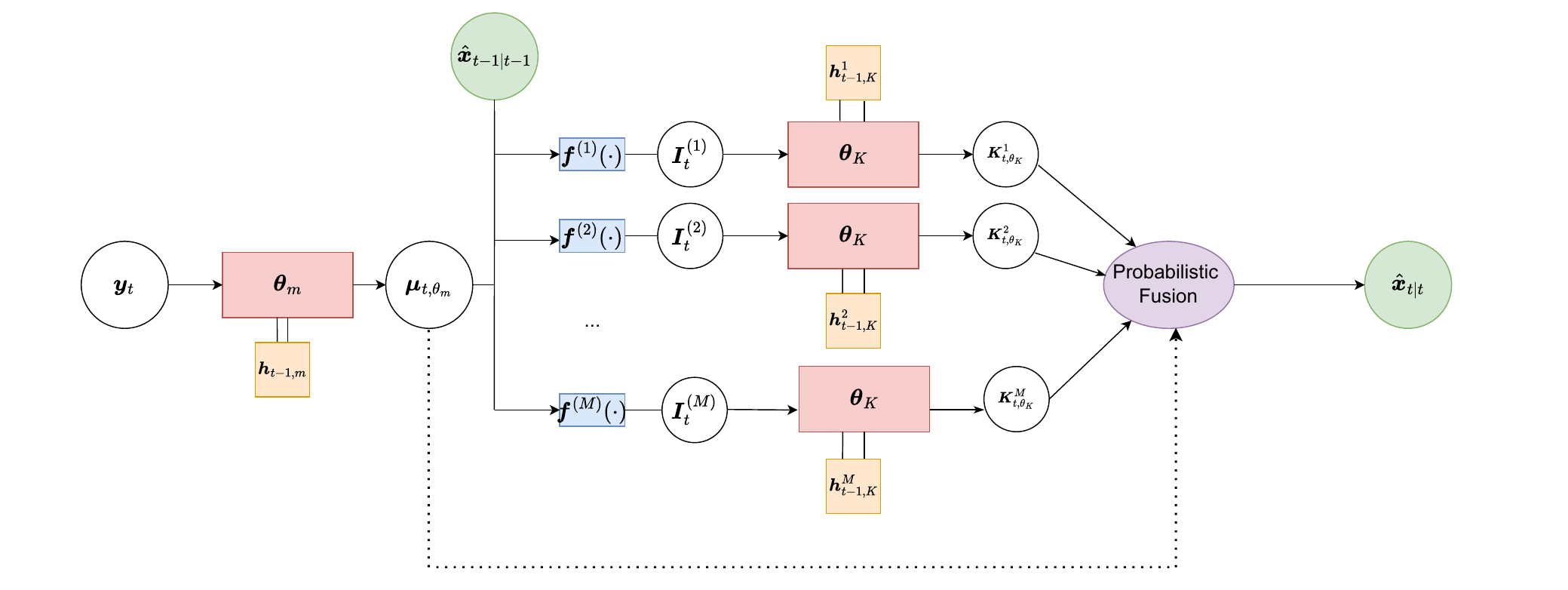}}\caption{The proposed JMFNet architecture. The mode prediction NN $\boldsymbol{\theta}_m$ processes the most recent $\boldsymbol{y}_t$ in order to assign probabilities to each mode. For each mode $j$, the prediction step is performed, and the input vectors $\boldsymbol{I}_t^{(j)}$ are constructed and passed at the mode-augmented KalmanNet $\boldsymbol{\theta}_K$. The resulting KG approximation can then construct the new state estimates $\mathbf{x}_{t|t}^{(j)}$. Finally, the resulting outputs are fused using probabilities $\boldsymbol{\mu}_{t,\boldsymbol{\theta}_m}$ as weights, resulting in the final state estimate.}\label{fig:our_filter}
\end{figure*}
All in all,  at each time instance $t$, our algorithm performs the following steps: \begin{enumerate}
    \item First, the observation $\mathbf{y}_t$ is passed to NN $\boldsymbol{\theta}_m$ in order to generate the mode probabilities $\boldsymbol{\mu}^{(j)}_{t,\boldsymbol{\theta}_m}$ $\forall j$.
    \item For each mode $j$, the prediction step remains the same as in the typical IMM (i.e., using $\mathbf{F}^{(j)}$ or $\mathbf{f}^{(j)}(\cdot)$ as in~(13a) or~(\ref{eq:IMM-nonLinearUpdates})).
    \item The update step is modified: the output $\mathbf{K}_{t,\boldsymbol{\theta}_K}^j$ of the second NN is used to update the state estimate as:
    \begin{equation}\label{eq:ours_individ_update}
        \hat{\mathbf{x}}^{(j)}_{t|t} \gets \hat{\mathbf{x}}_{t|t-1}^{(j)}+\mathbf{K}_{t,\boldsymbol{\theta}_K}^j (\mathbf{y}_t -\hat{\mathbf{y}}_{t|t-1}^{(j)}).
    \end{equation}
    \item In the fusion step, the final state estimate is computed according to expression: \begin{equation}\label{eq:NeuralFusion}
        \hat{\mathbf{x}}_{t|t} \leftarrow \sum_{j=1}^{M} \boldsymbol{\mu}_{t,\boldsymbol{\theta}_m}^{(j)}  \hat{\mathbf{x}}^{(j)}_{t|t}.
    \end{equation}
\end{enumerate}
\SetKwInput{KwData}{Input}
\begin{algorithm}[!t]
\caption{The Proposed JMFNet}\label{alg:filter_operation}  
\KwData{Models $\mathbf{f}^{j}(\cdot)$ and $\mathbf{h}^{j}(\cdot)$ $\forall j \in \{1,2,\ldots,M\}$; trajectory horizon $T$; NNs $\boldsymbol{\theta}_m$ and $\boldsymbol{\theta}_K$}
Initialize hidden states $\mathbf{h}_{0,m}$ and $\mathbf{h}_{0,K}$ for both RNNs.\\
Set $\mathcal{L}=0.0$.\\  
\For{$t=1,2,\ldots,T$}{
1) Observe $\boldsymbol{y}_t$.\\
2) Compute the mode probabilities $\boldsymbol{\mu}_{t,\boldsymbol{\theta}_m}$ using the $\hspace*{0.21cm}$ mode prediction NN $\boldsymbol{\theta}_m$.\\  
\For{$j=1,2,\ldots,M$}{
3) Compute $\hat{\mathbf{x}}^{(j)}_{t|t-1}, \hat{\mathbf{y}}_{t|t-1}^{(j)}$ via~(\ref{eq:IMM-nonLinearUpdates}) using  $\hspace*{0.21cm}\hat{\mathbf{x}}_{t-1|t-1}$.\\
4) Calculate input $\mathbf{I}_t^{(j)}$ using~(\ref{eq:InputType1}).\\ 
5) Compute $\mathbf{K}_{t,\boldsymbol{\theta}_K}^j$ using the mode-augmented $\hspace*{0.21cm}$KG estimator $\boldsymbol{\theta}_K$. \\ 
6) Predict $\mathbf{x}_{t|t}^{(j)}$ via~(\ref{eq:ours_individ_update}).
}
7) Fuse outputs into $\hat{\mathbf{x}}_{t|t}$ using~(\ref{eq:NeuralFusion}). \\
8) Update $\mathcal{L} \gets \mathcal{L}+||\mathbf{x}_t - \hat{\mathbf{x}}_{t|t}^{(j)}||^2$.
}
Return $\mathcal{L}$.
\end{algorithm}

This newly fused state estimate will be used for the prediction step in the next time instance $t+1$.
The overall operation of the proposed filter is described in Algorithm~\ref{alg:filter_operation} and visualized in Fig.~\ref{fig:our_filter}.

\paragraph*{Complexity and Stability Improvements over IMM}
A traditional IMM filter  scales approximately as $O(M s ^3)$ due  to the need to run in parallel $M$ KF updates and covariance matrix propagations. In contrast, JMFNet completely skips covariance computation, leading to a complexity of $O(M (s^2+t_n))$, where $O(M s^2)$ is required for the Kalman prediction step and $t_n$ is the forward inference time complexity of the trained NNs.  With the aid of modern GPUs, lighweight RNNs, like the ones used in Section~\ref{sec:experiments}, can perform inference decisions very quickly; this implies that we can effectively assume that $t_n \in \mathcal{O}(1)$. Furthermore, classical IMM and EKF-based variants require repeated matrix inversions to update  covariance matrices which run the risk of becoming ill-conditioned on high dimensional and complicated systems, especially when combined with  Jacobian-based linearization.
\subsection{ALS-Based Training}
To train the two RNNs, we employ an ALS optimization scheme on the following loss function. Let $D$ denote the total number of training sequences, and let $T^{(d)}$ be the length of sequence $d$. Our objective is to minimize the discrepancy between the true states and the predicted filtered ones, with the latter computed 
using the outputs of both $\boldsymbol{\theta}_m$ and $\boldsymbol{\theta}_K$ according
to~(\ref{eq:NeuralFusion}) for each time instance $t$. In mathematical terms, the total loss is defined as follows:
\begin{equation}
     \label{eq:LossFunct}
     \mathcal{L}(\boldsymbol{\theta}_m,\boldsymbol{\theta}_K)\triangleq \sum_{d=1}^{D} \sum_{t=1}^{T^{(d)}} \left\|\mathbf{x}_t - \hat{\mathbf{x}}_{t|t}\right\|^2. 
 \end{equation}

The ALS procedure alternates between updating the mode prediction NN parameters $\boldsymbol{\theta}_m$ and the KG estimation parameters $\boldsymbol{\theta}_K$, holding one fixed while optimizing the other with a stochastic gradient descent variant, such as Adam~\cite{Adam}. This decoupling allows each network to specialize, as follows: the mode prediction NN focuses on inferring the most likely mode sequence given the observations, while the mode-informed KalmanNet learns to best estimate the continuous state conditioned on the current mode. By iteratively optimizing each component, the model converges to a joint solution that effectively captures both the discrete and continuous aspects of the underlying system dynamics. It is noted that the use of ALS has been motivated by the coupled nature of the learning problem. The prediction of the discrete mode sequence and the estimation of the continuous state are interdependent. Thus, jointly optimizing both components is challenging due to the circular dependency, i.e., errors in mode prediction can degrade state estimation, and vice versa.

Focusing on the loss in~\eqref{eq:LossFunct} for a single instance and fixed $\boldsymbol{\theta}_K$, $\mathcal{L}_t\triangleq||\mathbf{x}_t-\hat{\mathbf{x}}_{t|t}||^2$, differentiation w.r.t. $\boldsymbol{\theta}_m$'s output gives: 
\begin{equation}
    \label{eq:derivForMode}
    \frac{d \mathcal{L}_t}{ d \boldsymbol{\mu}_{t,\boldsymbol{\theta}_m}^{(j)}}=-2\left(\mathbf{x}_t-\hat{\mathbf{x}}_{t|t}\right)\frac{ d\hat{\mathbf{x}} _{t|t}}{ d \boldsymbol{\mu}_{t,\boldsymbol{\theta}_m}^{(j)}}= -2\left(\mathbf{x}_t-\hat{\mathbf{x}}_{t|t}\right)\hat{\mathbf{x}}_{t|t}^{(j)},
\end{equation}
where $j=1,2,\ldots,M$.
This gradient computation indicates that  the end-to-end Least Squares (LS) loss function is differentiable w.r.t. the weights of the mode prediction NN, hence, gradient-driven optimizers can be directly applied without needing to construct any ``intermediate'' losses.
Likewise, for fixed $\boldsymbol{\theta}_m$, the gradient is computed as follows:
\begin{align}
    \label{eq:derivForKG}
    \frac{d \mathcal{L}_t}{ d \mathbf{K}^j_{t,\boldsymbol{\theta_K}}}&=-2\left(\mathbf{x}_t-\hat{\mathbf{x}}_{t|t}\right) \frac{ d\hat{\mathbf{x}} _{t|t}}{ d \mathbf{K}^j_{t,\boldsymbol{\theta_K}} } \nonumber \\
    &= -2\left(\mathbf{x}_t-\hat{\mathbf{x}}_{t|t}\right) \boldsymbol{\mu}_{t,\boldsymbol{\theta}_m}^{(j)} \left(\mathbf{y}_t-\hat{\mathbf{y}}_{t|t-1}^{(j)}\right),
\end{align}
implying that the end-to-end LS loss is differentiable w.r.t. $\boldsymbol{\theta}_K$'s weights, and directly promotes KG estimation.

In practice, mini-batch versions of~(\ref{eq:LossFunct}) will be optimized. In particular, the dataset of $D$ trajectories is split into $N_B$ batches of size $B$. The $b$-th batch (with $b=1,2,\ldots,B$) corresponds to the trajectory indices $d_{b,1},d_{b,2},\ldots,d_{b,B}$, and the mini-batch loss is given by:
\begin{equation}
    \mathcal{L}(\boldsymbol{\theta}_m,\boldsymbol{\theta}_K)_b = \sum_{d=d_{b,1}}^{d_{b,B}} \sum_{t=1}^{T^{(d)}} \left\| \boldsymbol{x}_t -\hat{\boldsymbol{x}}_{t|t}\right\|^2.
\end{equation}

As it will be experimentally demonstrated in the numerical results' Section~\ref{sec:experiments} that follows, the adopted ALS procedure is robust to initialization seeds, small changes in the learning rate, and in the batch size $B$
. 
A high level overview of the proposed training scheme is provided in Algorithm~(\ref{alg:ALS}).
\begin{algorithm}[!t]
\caption{ALS-Based Training of JMFNet}\label{alg:ALS} 
Initialize $\boldsymbol{\theta}_m$ and $\boldsymbol{\theta}_K$.\\
\For{epoch$=1,2,\ldots,E$}{
\textit{Optimize mode prediction with fixed KalmanNet:}\\
\For{$b=1,2,\ldots,N_B$}
{
1) Filter the trajectories using Algorithm~(\ref{alg:filter_operation}).\\ 
2) Optimize $\boldsymbol{\theta}_m$ using Adam on $L(\boldsymbol{\theta}_m,\boldsymbol{\theta}_K)_b$.

}
\textit{Optimize KG estimator with fixed mode prediction.}\\
\For{$b=1,2,\ldots,N_B$}
{

3) Filter the trajectories using Algorithm~(\ref{alg:filter_operation}).\\ 
4) Optimize $\boldsymbol{\theta}_K$ using Adam on $L(\boldsymbol{\theta}_m,\boldsymbol{\theta}_K)_b$.

}
}
\end{algorithm}
\subsection{Backpropagation Through Time for RNNs}
In order to compute the gradients required by Adam~\cite{Adam} (or related optimizers) for RNN architectures, such as the ones deployed in the proposed filter, BackPropagation Through Time (BPTT)~\cite{BBPT} is needed. The following two cases have been considered: \begin{itemize}
    \item[\textit{i})] Short-to-moderate length trajectories: in this case, typical BBPT is applied at the end of the episode.
    \item[\textit{ii})] Longer trajectories: truncated BPTT~\cite{truncBBPT} can be used, splitting the trajectories into shorter segments, which are shuffled and used during training.
\end{itemize}

\section{Numerical Results and Discussion}\label{sec:experiments}
In this section, we first discuss the parameters' setting for the proposed filtering framework, then detail the implemented benchmark schemes for comparison purposes, and finally present numerical experiments on various linear and nonlinear systems, including target tracking, pendulum angle tracking, Lorenz attractor dynamics, and a real-life dataset.

\subsection{Implementation of the Proposed Filter}
We have implemented JMFNet using the pytorch framework~\cite{pytorch}. The mode prediction NN was a GRU with two hidden layers of $64$ units each, whereas the mode-augmented KalmanNet began with a linear layer that mapped $\boldsymbol{I}_t^{(j)}$ to a hidden state of dimension $4 \times n_I$, where $n_I$ was the input dimension. This layer was followed by a single-layer GRU with hidden size $6 \times n_I$ and a final linear layer that provided KG as output. All hyperparameters for training the RNNs are detailed in Table~\ref{tab:ALS_hyperparameters}. Experiments were conducted on an NVIDIA GeForce RTX $3080$ GPU with $32$~GB of memory. Throughout this section, the results for our method have been averaged over $10$ different initialization seeds.

\begin{table}[!t] 
    \centering
    \begin{tabular}{|c|c|}
    \hline
         Parameter&Value  \\
         \hline
         \hline
         Learning rate& $5\times10^{-4}$\\
         \hline
				 Training set size & $40000$\\
				\hline
         Validation set size & $10000$\\
				\hline
         Batch size ($B$) & $16$\\ 
				\hline
         Gradient clip norm & $2.5$\\ 
				\hline
         Training iterations ($E$) & $50$\\
				\hline
    \end{tabular}
    \caption{Hyperparameter values for both proposed RNNs.}
    \label{tab:ALS_hyperparameters}
\end{table}
\subsection{Benchmarks}
The following baselines have been implemented and tested for SS estimation comparison purposes.
 \begin{enumerate}
    \item IMM algorithm: The one described in Section~\ref{subsec:IMM}; when the underlying dynamics are linear, it uses KF, otherwise EKFs.
      \item Particle Filters (PFs): Sequential Monte Carlo estimators like PFs~\cite{PF1,PF2} are very popular MB estimators and have been successfully used in linear systems with switches (see, e.g., \cite{PFKrisn}). We have implemented a standard bootstrap PF tailored to the switching motion model described in Section~\ref{subsec:jump_system_model}. The number of particles was set to $200$.
    \item Switch-Agnostic KalmanNet: This two-architecture framework is intended for comparisons with the KalmanNet. This filter assumes that the first out of $M$ modes are correct. 
    \item MF RNNs: Two RNNs, LSTM and GRU, that use their hidden memory to map $\boldsymbol{y}_{1:t}$ to $\boldsymbol{x}_t$ have been considered. Both networks had similar architectures and training parameters with the proposed mode-augmented KalmanNet. Since they were not considered to have access to model dynamics, the sizes of all hidden layers and the training iterations were doubled. 
      \item MF Transformer: A standard transformer architecture~\cite{attentionIsAllYouNeed} trained similarly to the MF RNNs has been used. It processed full historical observations at each time instance using causal self-attention. The model employed a $2$-layer transformer encoder with $8$ attention heads of $64$ hidden units. Sinusoidal positional encodings preserved temporal order, while layer normalization was used to stabilize training. We have also used a dropout rate of $0.1$ to prevent overfitting. An NN with $2$ ReLU-activated layers of $256$ units was used to construct the final output. For each simulation, the learning rate varied from $0.00001$ to $0.001$. We have noticed that transformers require far more data to train, and, thus, take longer to converge, so we increased the training set size to $100000$ and the number of optimization iterations to $250$.
			
It is noted that the considered transformer architecture was deployed to capture long-range mode transitions without constraints, though at a higher computational costs. In particular, the cost was much higher than that of the proposed MB DL filter and of MF RNNs, making it unsuitable for certain signal processing tasks. In addition, the increased training time and size of the model made it hard to perform rigorous hyperparameter optimization. In fact, similar to our focus on state estimation in JMSs, limitations of off-the-self transformers for time series are well established~\cite{chen2023transformers}. For our benchmarking purposes, we have considered herein a ``vanilla'' transformer, but attention-based filters constitute an important direction for future work.
\end{enumerate}

\subsection{Results for 2D Target Tracking}
A two-dimensional (2D) target tracking application has been considered, where the target's true state was a $4$-dimensional vector $\boldsymbol{x}_t=[p_x,v_x,p_y,v_y]$ with $p_x$ and $p_y$ being the 2D position values and $v_x$ $v_y$ their respective velocities. The target alternates between different kinematic laws, and the proposed filter was deployed to accurately infer its position and velocity information through noisy position observations.
\paragraph{Linear Motion}
The following fixed transition matrix was considered:
\begin{equation}
    \boldsymbol{\Pi}= \begin{bmatrix}
        0.9 & 0.1 \\
        0.2 & 0.8
    \end{bmatrix},
\end{equation}
according to which the target switched between $M=2$ modes:
Constant Velocity (CV) and Constant Turn (CT). For the former mode, the state update matrix was set as:
\begin{equation}
    \mathbf{F}_{\rm CV}=\begin{bmatrix}
        1 & dt & 0 &0 \\
        0 & 1 & 0 & 0 \\
        0 & 0 & 1 & dt \\
        0 & 0 & 0 & 1
    \end{bmatrix}
\end{equation}
with $dt=1$~sec being the interval between two subsequent observations. For the CT mode, it was assumed that the target is turning with a constant angular velocity $\omega=0.1$~rad/sec, and the transition matrix was set to: 
\begin{equation}
    \mathbf{F}_{\rm CT}= \begin{bmatrix}
        1 & \sin\omega/\omega & 0 & -(1-\cos \omega)/\omega \\
		0 & \cos \omega &        0 & -\sin \omega \\ 
	0 & (1-\cos\omega)/\omega &  1 & \sin \omega/\omega \\
		0 &  \sin \omega &   0 &   \cos\omega
    \end{bmatrix}.
\end{equation}

\begin{figure*}[!t] 
    \centering  
    \begin{subfigure}[b]{0.32\textwidth}
        \centering
        \includegraphics[width=\textwidth]{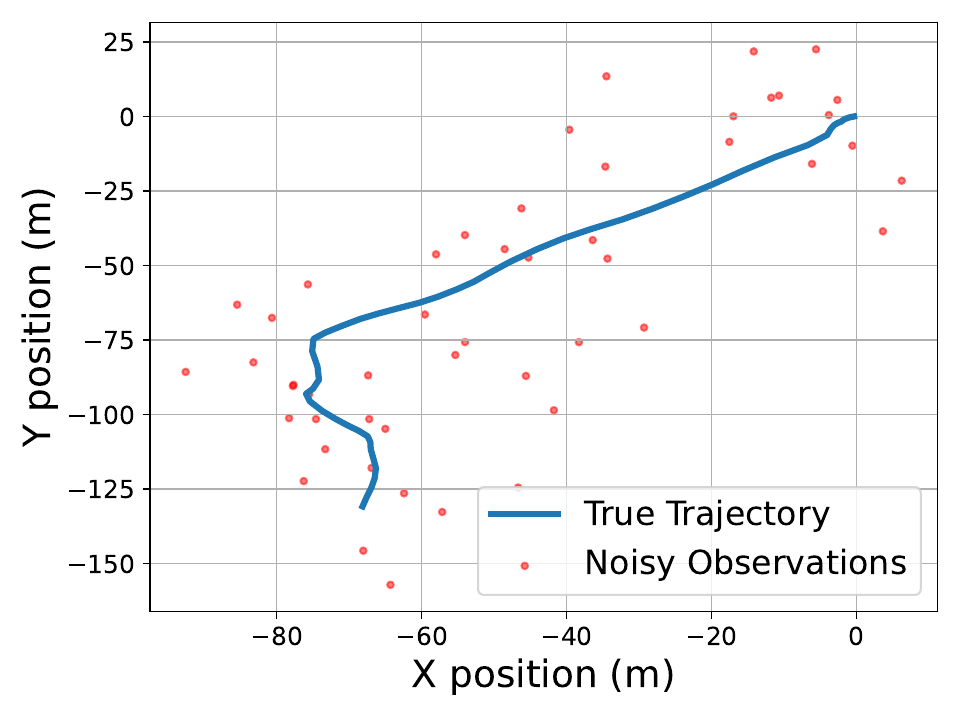}
        \caption{Linear with $M=2$, $T=50$.}
        \label{fig:exampleTrajSmall}
    \end{subfigure}
    \begin{subfigure}[b]{0.32\textwidth}
        \centering
        \includegraphics[width=\textwidth]{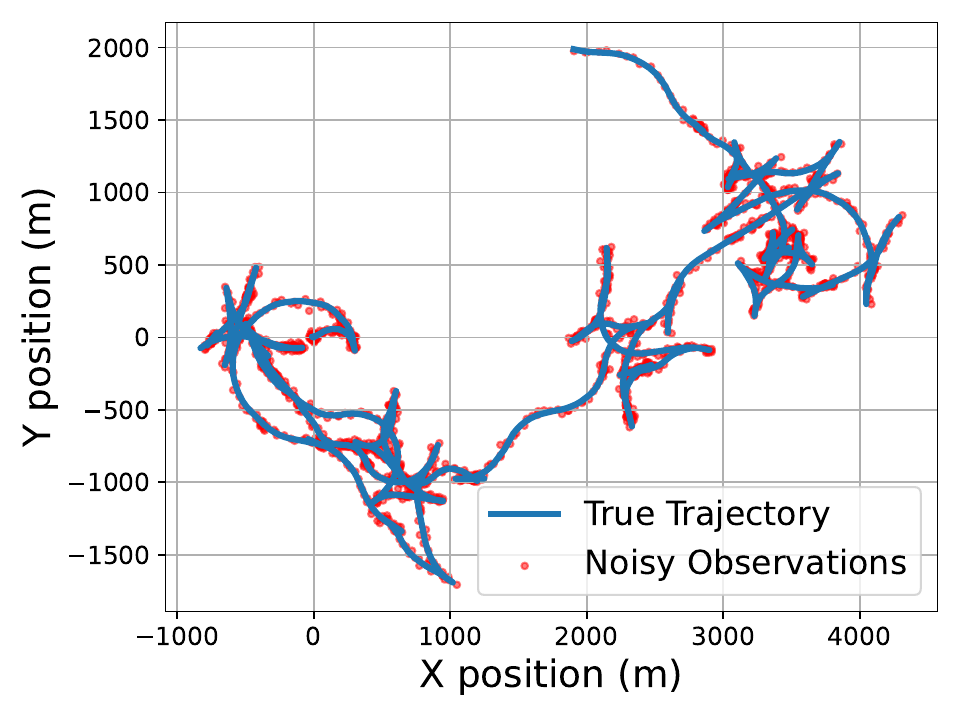} 
        \caption{Linear with $M=4$, $T=2000$.}
        \label{fig:exampleTraj4}
    \end{subfigure}
    \begin{subfigure}[b]{0.32\textwidth}
        \centering
        \includegraphics[width=\textwidth]{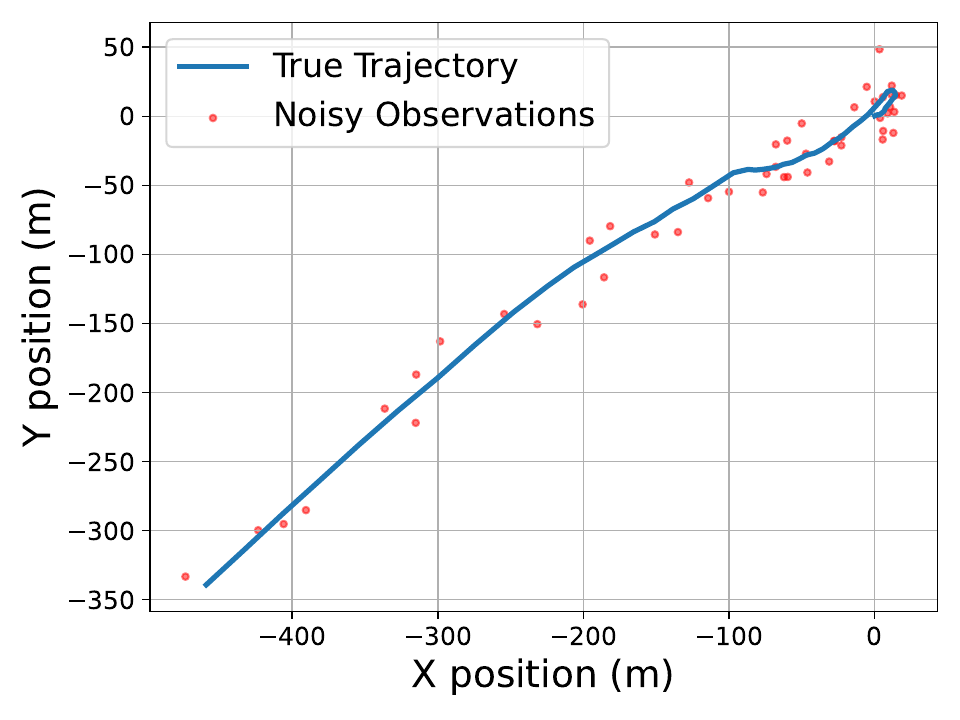} 
        \caption{Quadratic, $T=50$.}
        \label{fig:exampleTrajQuad}
    \end{subfigure}
    \caption{Example trajectories for the considered 2D target tracking application with different noise levels and of different time horizons, considering $q_{\rm CV}=0.5$, $q_{\rm CT}=2$, and $r_{\rm CT}=r_{\rm CV}=5$.}  
    \label{fig:example_trajectories_target_trackin}
\end{figure*}

The goal of the proposed filtering approach is to accurately estimate the target's state by only observing location information, i.e., in \eqref{Lds-y} for both modes: 
\begin{equation}
    \mathbf{H} = \begin{bmatrix}
        1 & 0 & 0 & 0\\
        0 & 0 & 1 &0
    \end{bmatrix}.
\end{equation}
Unless stated otherwise, we have considered zero-mean Gaussian noise, with the following covariance structure: 
\begin{equation}\label{eq:noise_target_tracking}
    \mathbf{Q}_{m}= q_m \mathbf{I}_4 \quad \text{and} \quad \mathbf{R}_m=r_m \mathbf{I}_2, 
\end{equation}
where $m$ is a string taking the value CV or CT. We have simulated various 2D target trajectories under different motion models setting $q_{\rm CV}=0.5$, $q_{\rm CT}=2$, and $r_{\rm CT}=r_{\rm CV}=5$, and considering the initial target position at the point $(0,0)$~m and its velocity for both axes to be $10$~m/sec. Figure~\ref{fig:example_trajectories_target_trackin} depicts two trajectories following the linear motion model, one for $M=2$ and the other for $M=4$ modes, and one trajectory for the quadratic motion model that will be described in the sequel. This figure includes both the true trajectories and their respective noisy measurements.
\begin{figure}[!t] 
        \centering
        \includegraphics[width=0.46\textwidth]{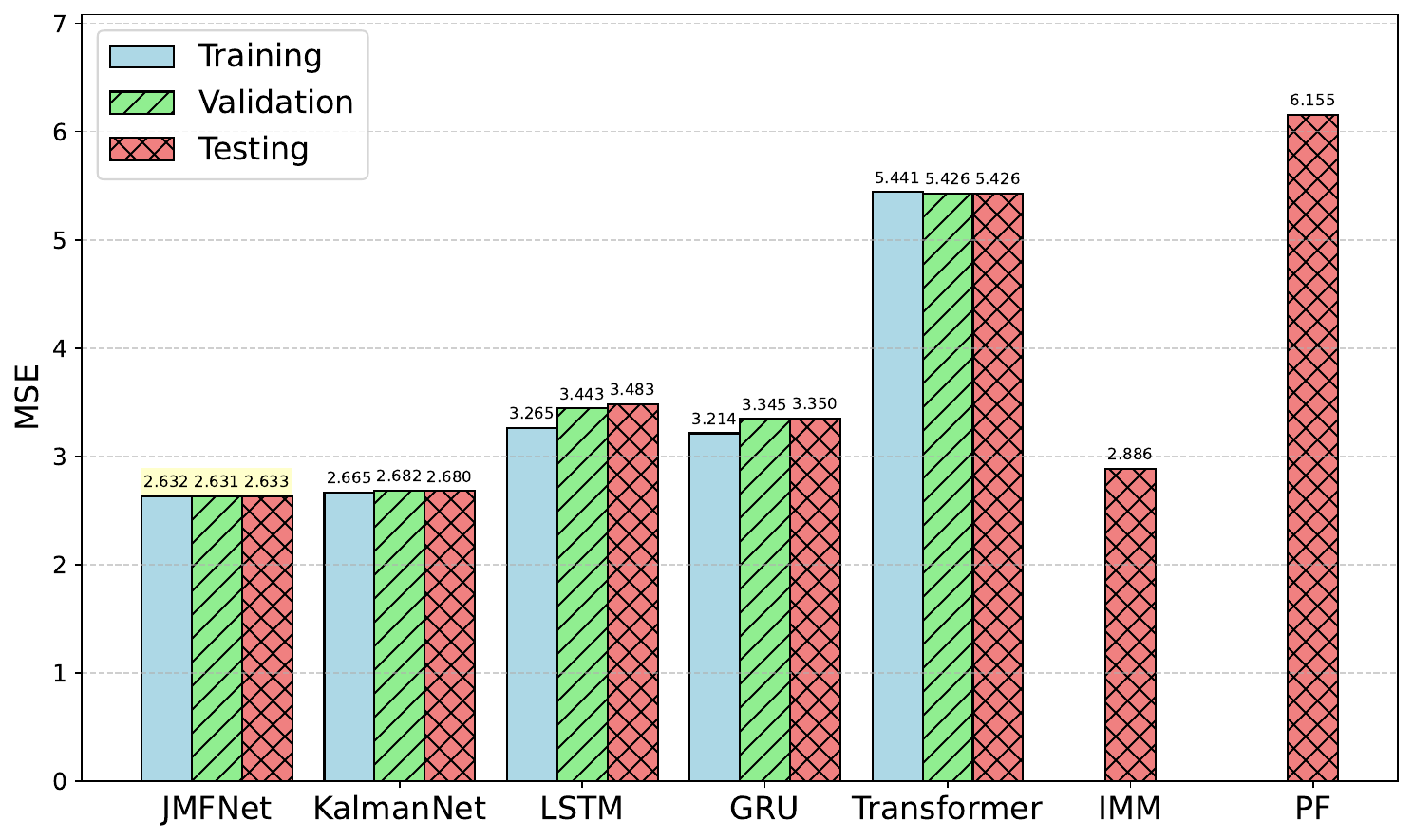} 
    \caption{State estimation scores for the linear 2D motion model with $M=2$ modes, considering a horizon length of $T=50$ and the Gaussian noise configuration of Fig.~\ref{fig:example_trajectories_target_trackin}.} 
    \label{fig:linear_target_simple}
\end{figure}

Focusing on the trajectories for the linear motion target with $M=2$ modes, and a horizon of $T=50$ samples, in  Fig.~\ref{fig:exampleTrajSmall},    Fig.~\ref{fig:linear_target_simple} depicts the state estimation scores, for both training, validation, and testing, for all considered benchmark schemes\footnote{The performance of non-learning benchmarks is included solely on the test set for two primary reasons: \textit{i}) their significant computational cost makes large-scale evaluation on $4\times10^4$ sequences impractical; and \textit{ii}) the conventional evaluation paradigm in classical MB signal processing is not concerned with dataset-wide generalization, but rather with achievable performance on specific, often illustrative, examples.}  and the proposed filtering framework. As observed, the proposed JMFNet approach  outperforms the purely MB and MF benchmarks. Surprisingly, KalmanNet achieves good performance despite ignoring the CT mode. The latter is actually the reason why our method outperforms KalmanNet by a small, yet notable margin. 
Moreover, it is shown that the large computationally expensive transformer and the PF (which is a computationally demanding estimator as well) perform worse than  the smaller RNNs and the IMM filter.

\begin{figure}[!t] 
    \centering
    \includegraphics[width=0.75\linewidth]{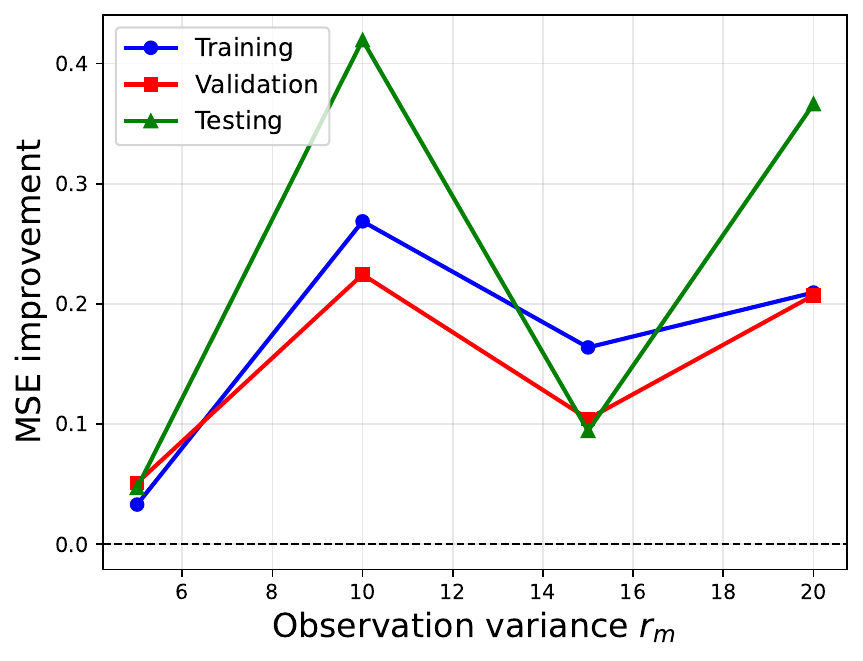}
    \caption{State estimation performance difference between the proposed filter and KalmanNet for the linear 2D motion model with $M=2$ modes versus different Gaussian noise levels $r_m$.}
    \label{fig:diffLinear}
\end{figure}

In addition, it is demonstrated in Fig.~\ref{fig:diffLinear} that the  improvement of JMFNet over the non-switching KalmanNet remains consistent for stronger observation noises. It will be showcased in the sequel that the proposed method can achieve a very big improvement over KalmanNet in more challenging switching patterns (e.g., chaotic systems) where the state and observation laws are substantially different between the modes. 

Figure~\ref{fig:exampleTraj4} depicts an example linear motion with $M=4$ modes, according to which the target can move backwards and turn with a negative angle. In particular, the constant velocity backward matrix was set as follows: 
\begin{equation}
    \mathbf{F}_{\rm CVB}=\begin{bmatrix}
        1 & -dt & 0 &0 \\
        0 & 1 & 0 & 0 \\
        0 & 0 & 1 & -dt \\
        0 & 0 & 0 & 1
    \end{bmatrix},
\end{equation}
whereas the constant inverse turn matrix was chosen to be as in $\mathbf{F}_{\rm CT}$ by just replacing $\omega$ with $-\omega$. 
In addition, the sequence length was set to $T=2000$ in order to investigate the ability of the proposed filter to capture long-term dependencies.

\begin{figure*}[!t]  
    \centering
    \begin{subfigure}[b]{0.3\textwidth}
        \scalebox{1.}{\includegraphics[width=\textwidth]{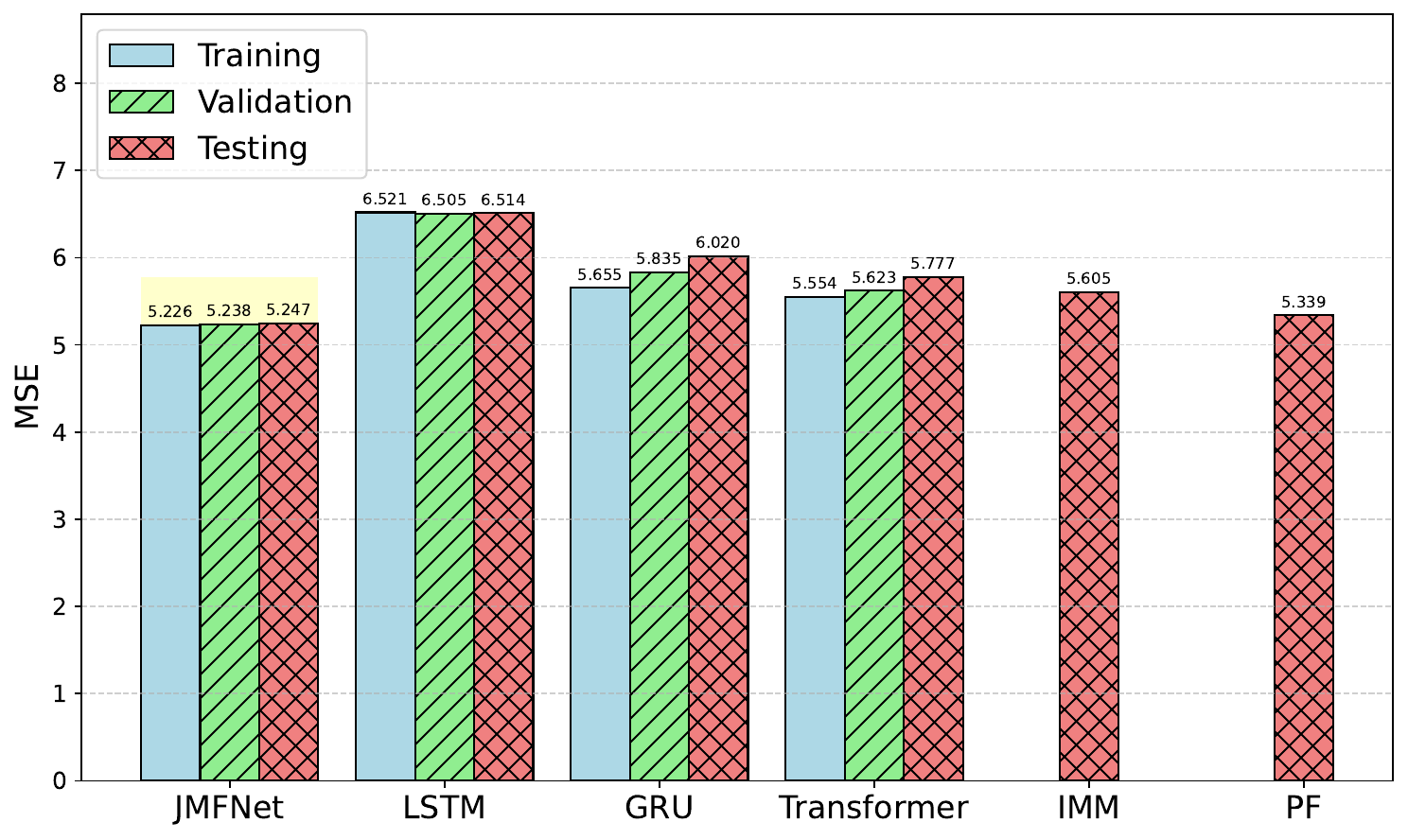}} 
        \caption{Gaussian noise.}
        \label{fig:sub1}
    \end{subfigure}
    \begin{subfigure}[b]{0.3\textwidth}
        \centering
        \scalebox{1}{\includegraphics[width=\textwidth]{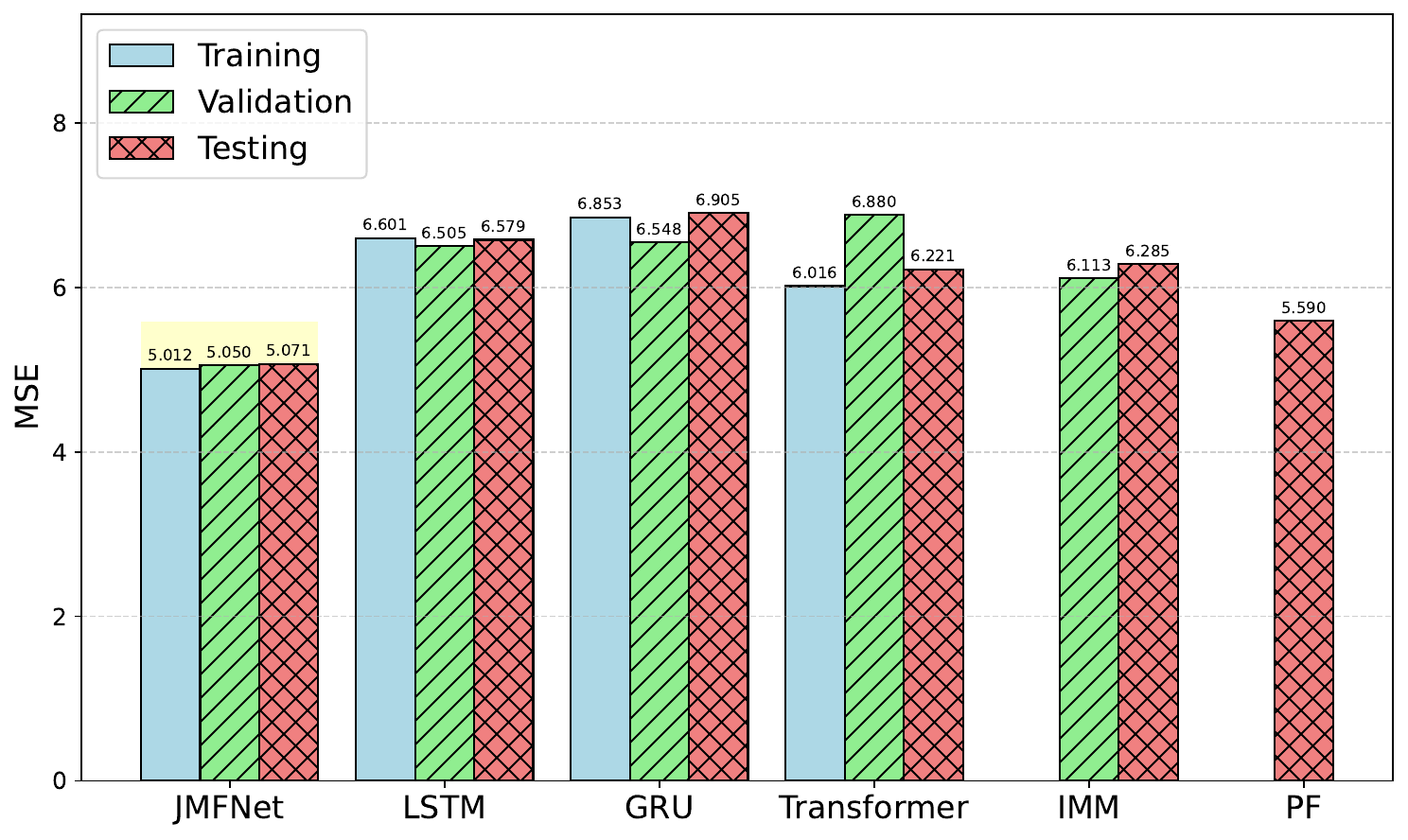}} 
        \caption{Laplacian noise.}
        \label{fig:sub2}
    \end{subfigure}
    \begin{subfigure}[b]{0.3\textwidth}
        \centering
        \scalebox{1}{\includegraphics[width=\textwidth]{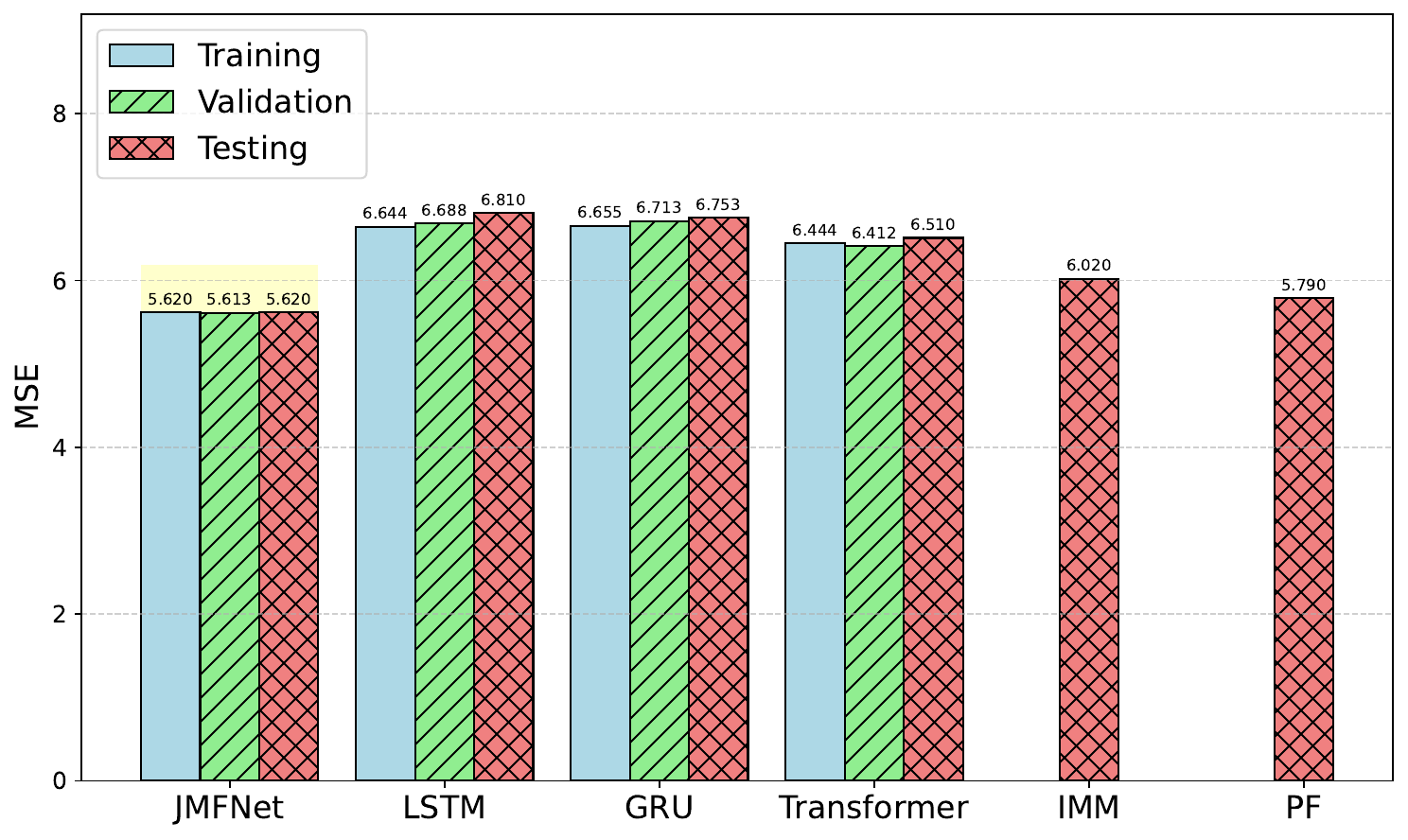}} 
        \caption{GMM noise.}
        \label{fig:sub2}
    \end{subfigure}
    \caption{State estimation scores for the linear 2D motion model with $M=4$ modes and a time horizon of $T=2000$, considering three different noise types.}
    \label{fig:lonng_sequences}
\end{figure*}

In Fig.~\ref{fig:lonng_sequences}, the motion model of Fig.~\ref{fig:exampleTraj4} was considered with the trajectories divided into segments of $20$ samples each (in order to apply truncated BPTT), including also training and validation scores with two additional observation noise models: the Laplacian model and the Gaussian Mixture Model (GMM). Their parameters were tuned to match the variance of the original Gaussian noise. To obtain more fine-grained discrete approximations of the  target motion, we have reduced the sampling interval to $dt = 0.01$ sec. It is demonstrated in the figure that the proposed filtering approach consistently outperforms all baselines. 

We also observed that the switch-agnostic KalmanNet,  that ignores backwards motion patterns, cannot filter these trajectories. This happens because for large $T$, assuming only constant motion (and no turns) the position estimates $p_x,p_y$ become very large, leading to large gradients and unstable training. It has been excluded from the demonstration due to very large MSE scores. that hindered presentation clarity.

\paragraph{Quadratic Motion}
We now turn our attention to nonlinear systems by introducing acceleration into the target's motion; the structure of the observations remains unchanged. In particular, two motion modes have been considered: constant and quadratic acceleration. In the former mode, the target's velocity was updated according to the standard kinematic equations:
\begin{subequations}\label{eq:linear_mode}
  \begin{align}
    v_x &\gets v_{x,t-1} + d_t a_x,  \\
    v_y &\gets v_{y,t-1} + d_t a_y,
  \end{align}
\end{subequations}
where the accelerations have been fixed to \( a_x = a_y = 1 \). For the latter mode, the acceleration was considered to vary quadratically with the target's position, as follows:
\begin{subequations}\label{eq:quadratic_mode}
\begin{align}
    a_x &\gets a x^2 + b x + c, \\
    a_y &\gets a y^2 + b y + c
\end{align}
\end{subequations}
with \( a \), \( b \), and \( c \) being predefined coefficients. This model introduces nonlinearity into the motion dynamics while maintaining a consistent observation model. 

\begin{figure}[!t] 
        \centering
        \scalebox{0.5}{\includegraphics[width=\textwidth]{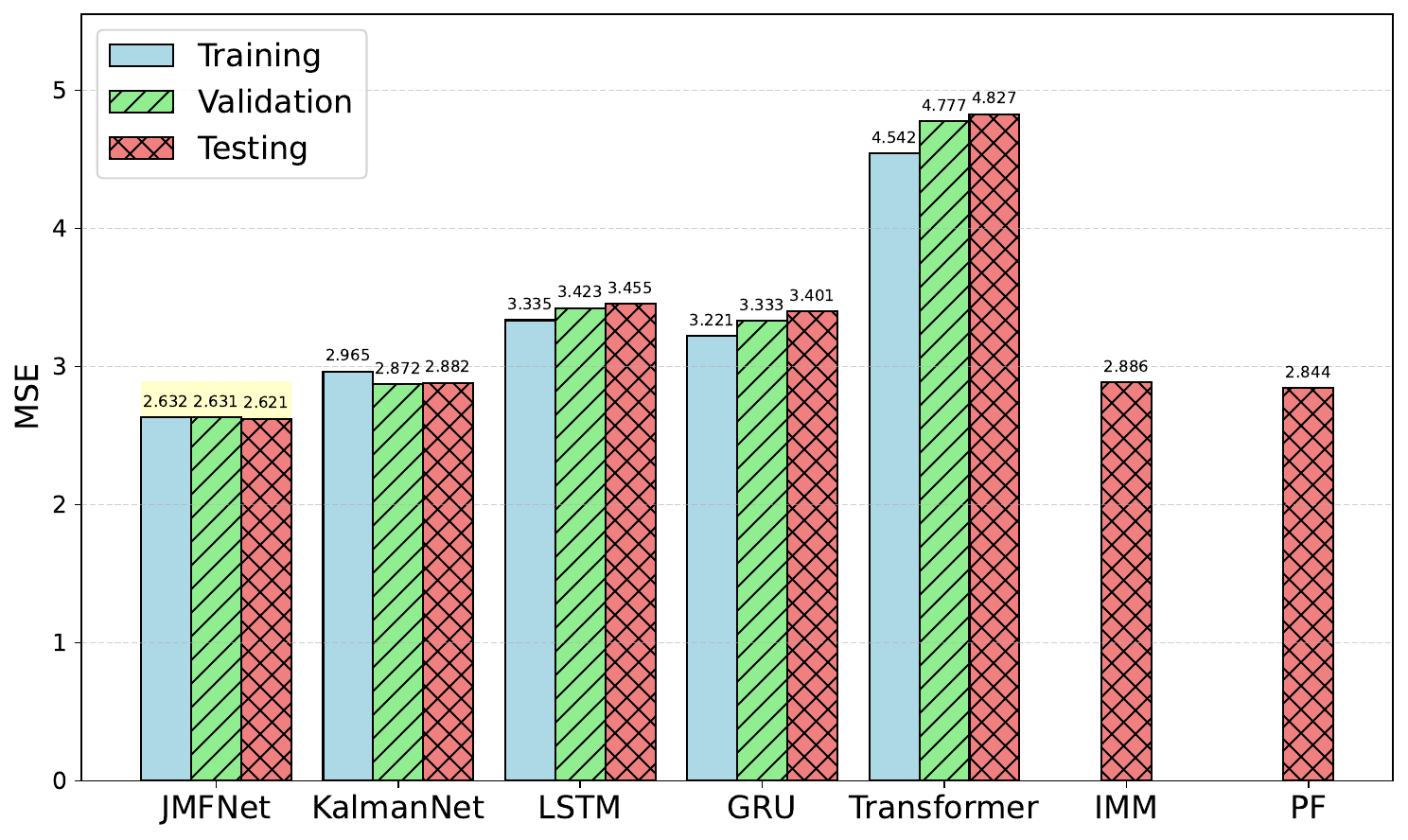}}
    \caption{State estimation scores for the quadratic 2D motion and a time horizon of $T=50$ samples, considering the Gaussian noise configuration of Fig.~\ref{fig:exampleTrajQuad}.}
    \label{fig:quadratic_results}
\end{figure}

Fig.~\ref{fig:quadratic_results} depicts the state estimation scores, for both training, validation, and testing, for all considered benchmark schemes using $T=50$ noisy observations via the Gaussian noise configuration of Fig.~\ref{fig:example_trajectories_target_trackin}. As depicted, JMFNet remains the superior approach, with KalmanNet also achieving satisfactory estimation performance. In particular, the proposed filter consistently achieves lower MSE than KalmanNet with a margin of $0.24$ to $0.33$ MSE units. 
(corresponding to a $8.3\%$ to $11.8\%$ reduction). While the absolute differences appear small, the improvement with the proposed approach is consistent across all datasets, underscoring its robustness and reliability over KalmanNet. It is noted that, in practical applications of JMSs, such as finance or surveillance, even small systematic improvements can have a disproportionately large impact on decision-making and long-term performance. In this application, model-agnostic learning-based filters perform significantly worse than traditional filters. 

\begin{figure}
    \centering
    \includegraphics[width=0.85\linewidth]{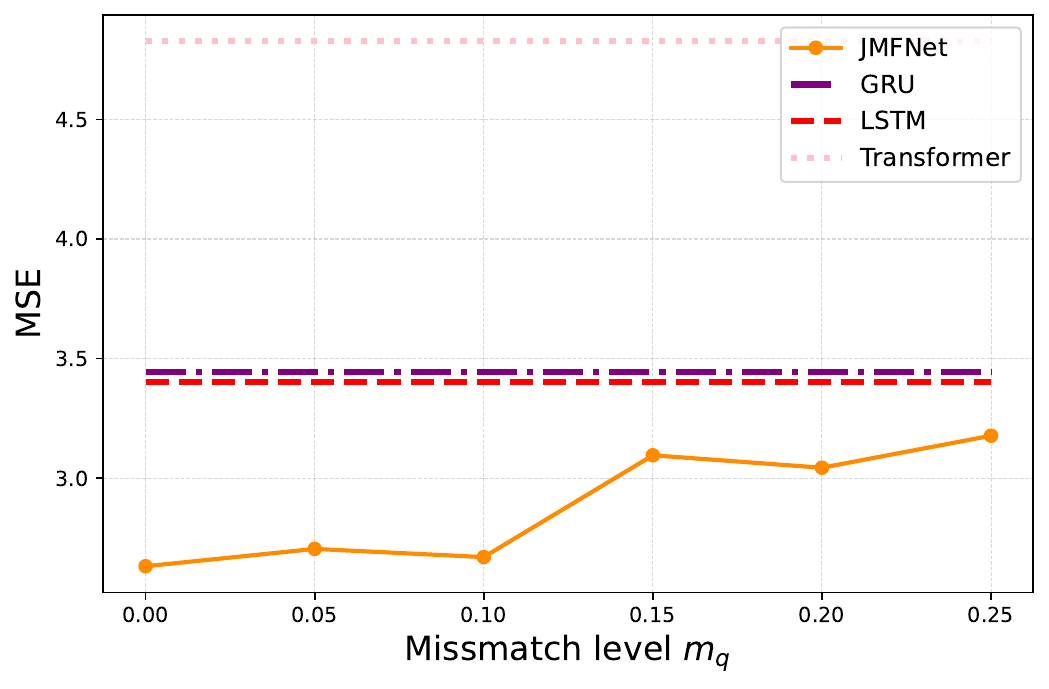}
    \caption{State estimation performance for the parameters' setting in Fig.~\ref{fig:quadratic_results} as a function of the different model mismatch levels $m_q$.}
    \label{fig:missmatch_results_quadratic_target} 
\end{figure}

\textbf{State Model Mismatch:}
Unlike MF RNNs, MB filters require precise specification of the underlying system model, making evaluation under incorrect model approximations crucial. To this end, the effectiveness of the proposed filter has been assessed under state update mismatch. We have particularly considered the target's 2D quadratic motion model in~(\ref{eq:linear_mode}) and~(\ref{eq:quadratic_mode}) including a bias term $m_q$ to the velocity and acceleration terms, causing the model to overestimate them both. In fact, a growing positional drift was introduced over time, which the RNNs of the proposed approached needed to correct. As shown in Fig.~\ref{fig:missmatch_results_quadratic_target}, JMFNet remains robust, outperforming MF baselines for different levels of $m_q$.

\begin{figure}
    \centering
    \includegraphics[width=0.85\linewidth]{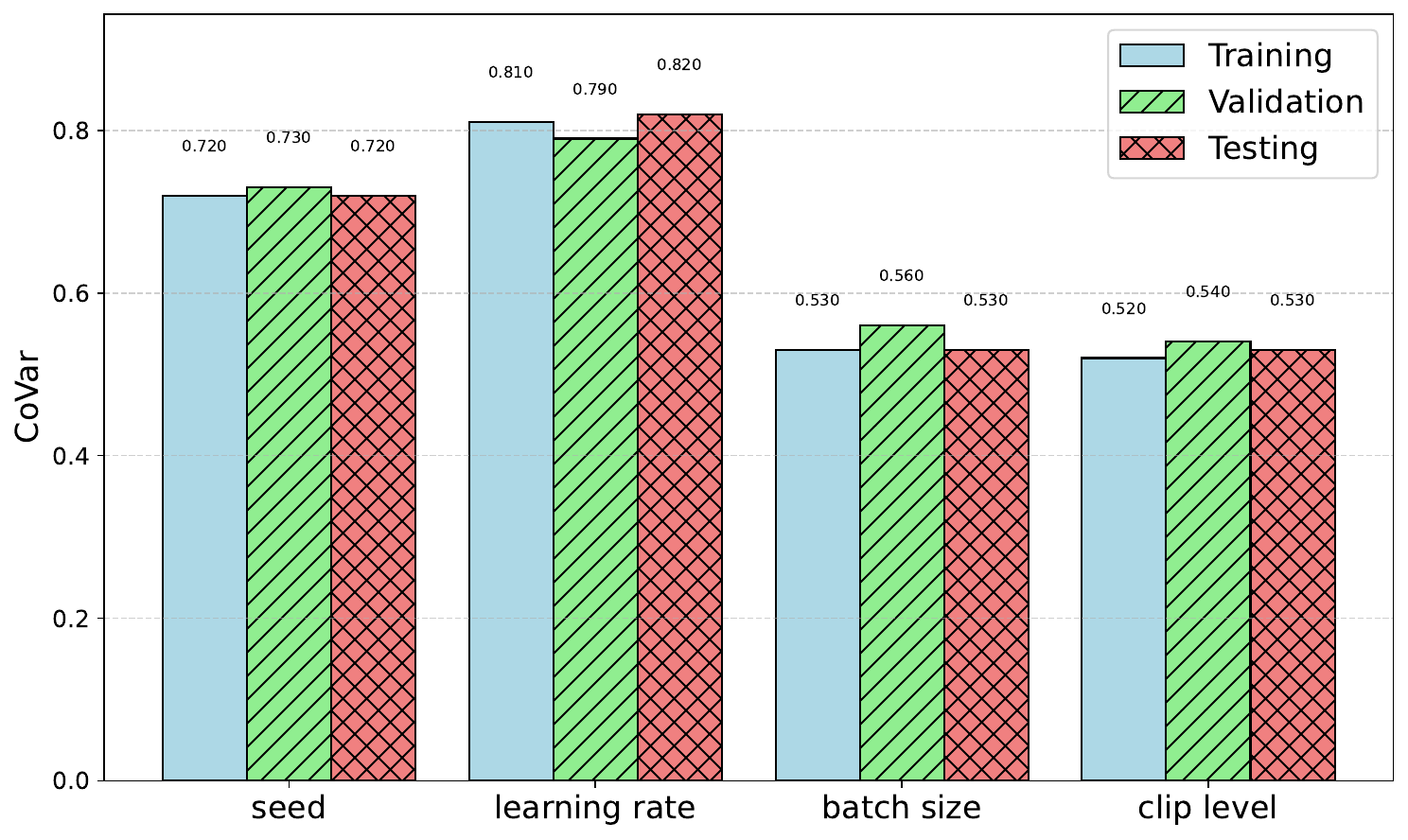}
    \caption{Sensitivity performance considering the parameters' setting in Fig.~\ref{fig:quadratic_results}.}
    \label{fig:quad_sensitivity}
\end{figure}

\textbf{Sensitivity Study:}
Training RNNs can be unstable, especially in long non-stationary trajectories. We have conducted sensitivity experiments to verify the stability of the proposed ALS-based training scheme as well as the generalization ability of the overall proposed filter. Considering the quadratic motion pattern, the role of the following parameters in the state estimation performance was investigated:
\begin{itemize}
    \item Initialization seed: training and testing were rerun with $25$ different initialization seeds.
    \item Learning rate: $10$ different learning rates, ranging from $10^{-4}$ to $10^{-3}$, were used.
    \item Batch size: The $B$ value was varied from $8$ to $64$.
    \item Gradient clipping level: Gradient clipping was used to prevent exploding gradients; $10$ different levels in $[1,5]$.
\end{itemize}
The Coefficients of Variation (CoVar) for each sensitivity experiment are demonstrated in Fig.~(\ref{fig:quad_sensitivity}). It can be observed that the proposed ALS-based training scheme is very robust to small changes in its hyperparameters, consistently achieving CoVar values less than $1$ implying that the standard deviation of the state estimation is less than $1\%$ of its mean value. 

\subsection{Results for Pendulum Angle Tracking}
A nonlinear pendulum system with switching dynamical modes has also been considered, described by the state vector $\mathbf{x}_t=\left[\theta_t,\dot{\theta}_t,\ddot{\theta}_t\right]$ at time instance $t$, 
where $\theta_t$ is the angle with respect to the vertical $y$-axis, $\dot{\theta}_t$ is the angular velocity, and $\ddot{\theta}_t$ is the acceleration. The angle was assumed to propagate according to the following Euler integration: 
\begin{subequations}
\begin{align}
    \dot{\theta}_{t+1} &\gets \dot{\theta}_t+\ddot{\theta}_t dt, \\
    \theta_{t+1} &\gets \theta_t+\dot{\theta}_{t}dt.
\end{align}
\end{subequations}
Finally, the system was assumed to switch between the following $M=4$ different modes: \begin{enumerate}
    \item Free undamped pendulum: For $g$ denoting the gravitational acceleration and $L=10$~m be the pendulum's length, it holds that: 
		\begin{equation}\label{eq:freePendulum}
        \ddot{\theta}_t= -\frac{g}{L} \sin{\theta}_t.
    \end{equation}
    \item Damped pendulum: The angular acceleration decreases according to a damping coefficient $\gamma=0.2$, i.e.: 
    \begin{equation}
        \ddot{\theta}_t= -\frac{g}{L} \sin{\theta}_t -\gamma \dot{\theta_t}.
    \end{equation}
    \item Driven Pendulum: An external harmonic force with amplitude $A_{\text{ext}}=1.0$~m and angular velocity $\omega_{\text{ext}}=2$~rad/sec is applied to the pendulum, influencing the acceleration pattern as:
    \begin{equation}
        \ddot{\theta}_t = -\frac{g}{L} \sin \theta_t +A_{\rm ext} \cos (\omega_{\rm ext} t).
    \end{equation}
    \item Random kicks: For $\xi_t$ being an impulsive noise term applied with probability $0.01$ at each discrete time instance $t$, the following model has been used:
		\begin{equation}
        \ddot{\theta}_t = -\frac{g}{L} \sin \theta_t +\xi_t.
    \end{equation}
\end{enumerate}
Each mode remained unchanged with probability $0.7$, and the probability of switching to all other modes was chosen as $0.1$. For each mode, noise of $\mathcal{N}(0,\sigma_{\text{pend}}\mathbf{I}_3)$ was added to $\mathbf{x}_t$. The goal of the SS estimation filter was to estimate the entire state $\mathbf{x}_t$ only from the noisy angle estimates $y_t= \theta_t+\mathcal{N}(0,10)$ (the observation noise $v_t^{(j)}$ was a univariate zero-mean Gaussian with variance of $10$ rad$^2$ for all $4$ modes) It is noted that, although the angle can be approximated from noisy sensors, knowledge of the other state components requires advanced measurement instruments like gyroscopes and accelerometers. 

\begin{figure*}[t]
    \centering
    \begin{subfigure}[b]{0.3\textwidth}
        \centering
        \includegraphics[width=\textwidth]{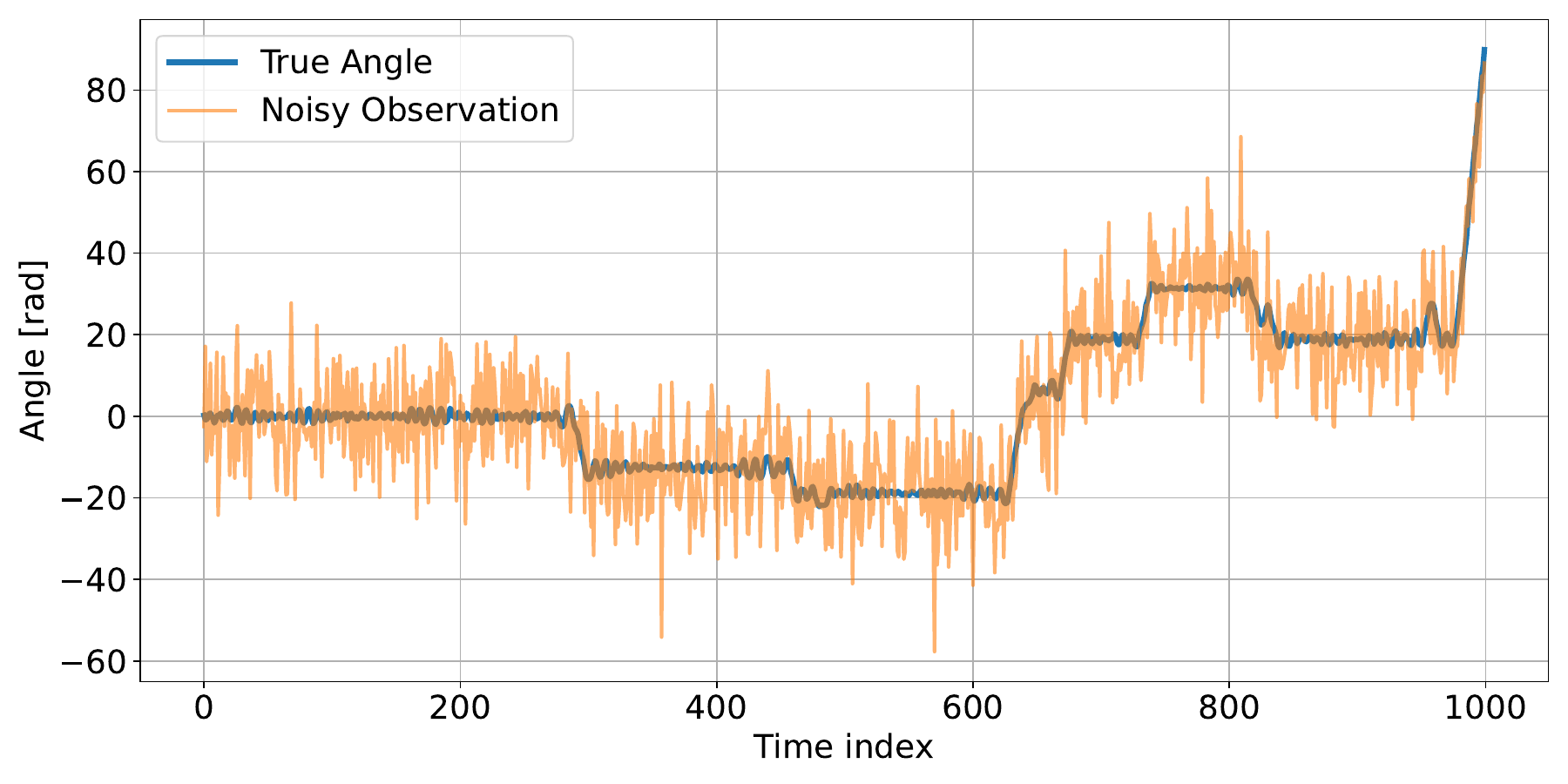}
        \caption{$\sigma_{\text{pend}}=0.01$.}
        \label{fig:exampleTrajPendulumLowNoise}
    \end{subfigure}
    \begin{subfigure}[b]{0.3\textwidth}
        \centering
        \includegraphics[width=\textwidth]{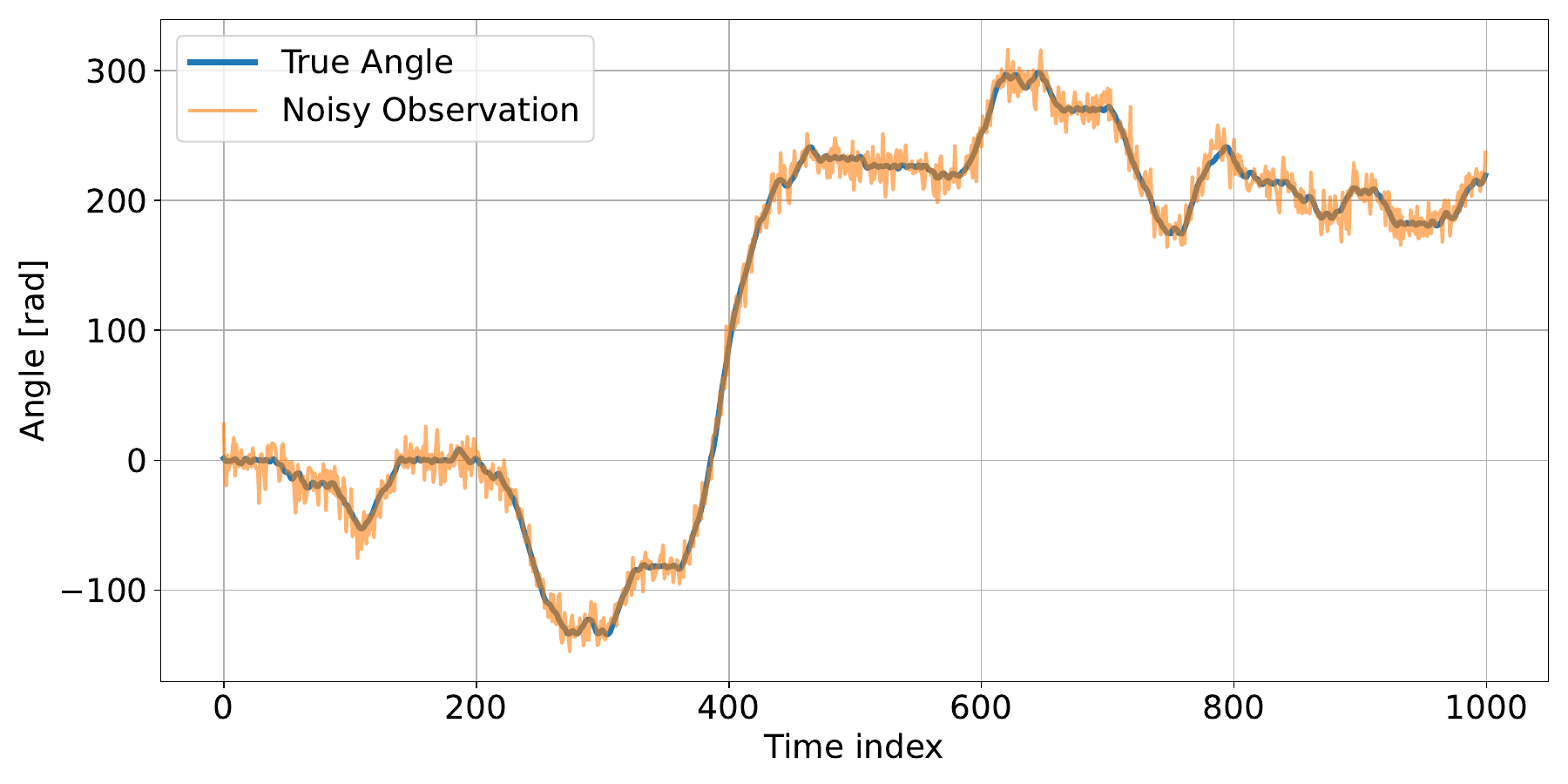} 
        \caption{$\sigma_{\text{pend}}=0.1$.}
        \label{fig:exampleTrajPendulumHighNoise}
    \end{subfigure}
    \begin{subfigure}[b]{0.3\textwidth}
        \centering
        \includegraphics[width=\textwidth]{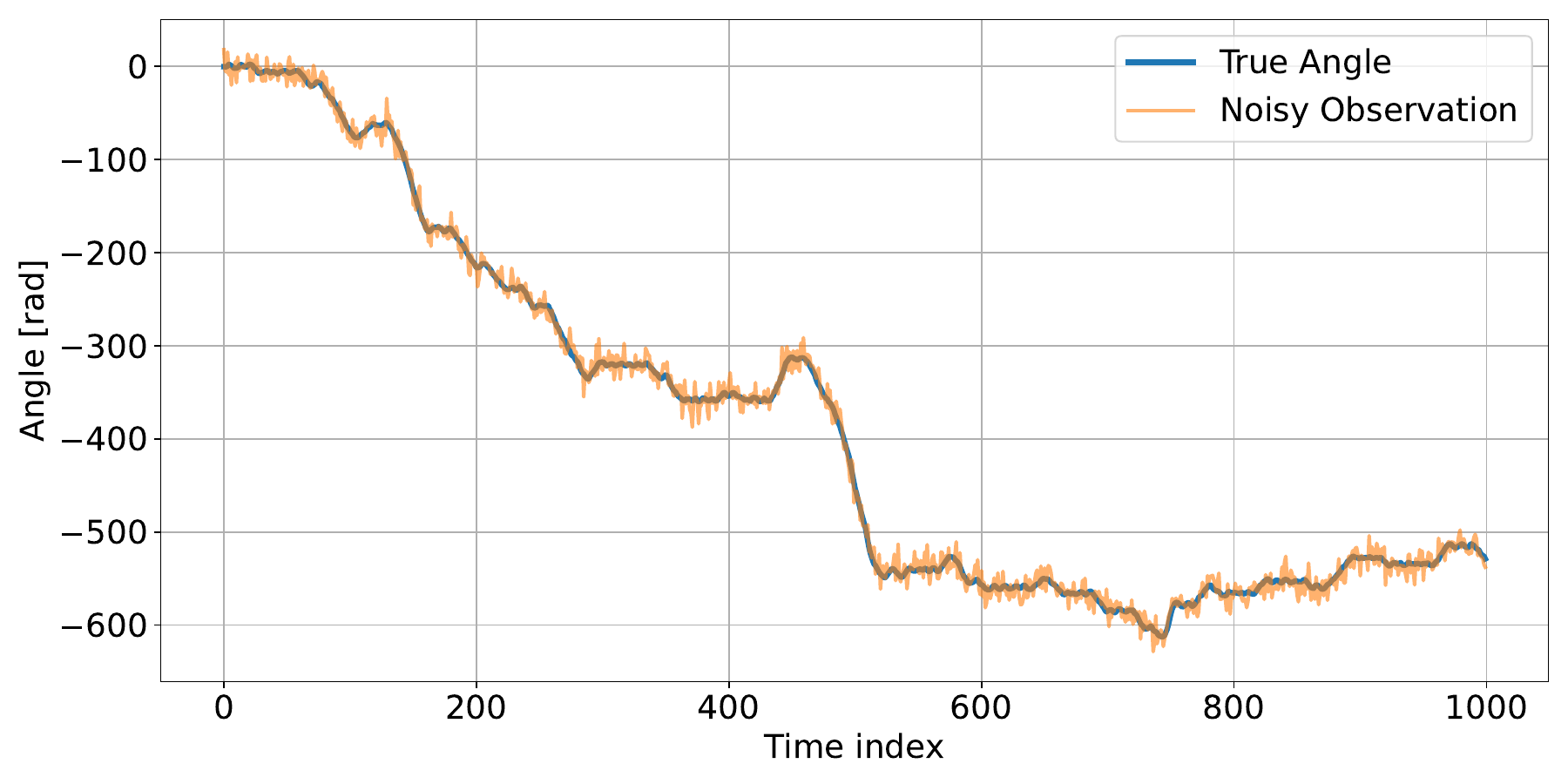} 
        \caption{$\sigma_{\text{pend}}=0.2$.}
        \label{fig:pendulumTrajHighNoise}
    \end{subfigure}
    \caption{Example angle  trajectories of the considered pendulum tracking application over a horizon of $T=1000$ samples for different values of the state noise level $\sigma_{\text{pend}}$.}
    \label{fig:pendulumExampleTraj}
\end{figure*}

\begin{figure}[]
        \centering 
        \includegraphics[width=0.4\textwidth]{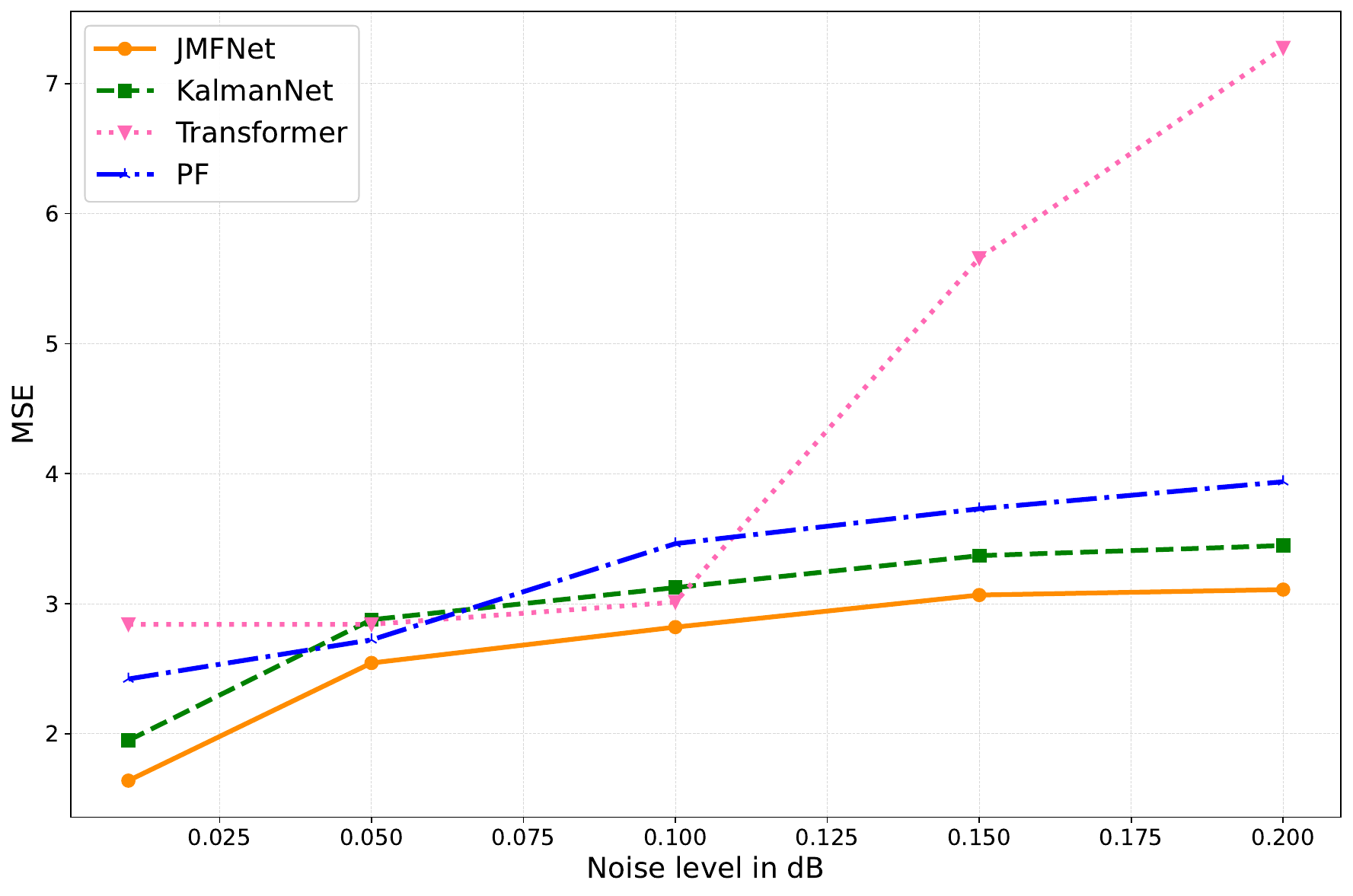} 
        \caption{Test set filtering performance versus the state variance $\sigma_{\text{pend}}$ for the pendulum tracking model in Fig. \ref{fig:pendulumExampleTraj}.} 
    \label{fig:pendulumMSE}
\end{figure}

We have considered a long horizon of $T=1000$ samples, which combined with the large observation noise, non-linearity, and the partial observations resulted in a very challenging state estimation problem. The noise parameter $\sigma_{\text{pend}}$ in the state evolution was varied from the very small value of $0.01$ to $0.2$. Example trajectories with different state variance levels are depicted in Fig.~\ref{fig:pendulumExampleTraj}. The performance results with all filters on the test set are reported in Fig.~\ref{fig:pendulumMSE}. The two MF RNNs and IMM performed poorly on this task and they have been excluded from the plot in order to enhance presentation quality. It can be seen that hybrid MB DL approaches outperform both MF filters and traditional MB filters. More importantly, it is shown that the proposed JMFNet filter consistently achieves smaller estimation error values than KalmanNet. 

\subsection{Results for the Lorenz Attractor}
Time-series datasets are often selected and preprocessed to aid forecasting. To this end, chaotic systems provide a popular and powerful testing toolkit for filtering, smoothing, and forecasting models~\cite{lorenzNeurIPS}. The state estimation performance of the proposed filter has been evaluated on synthetic trajectories simulated using the Lorenz attractor, which constitutes a three-dimensional (3D) solution to the continuous-time Lorenz equations. In particular, at each continuous time $\tau$, the state evolves according to:
\begin{equation}\label{eq:lorenzUpdate}
    \frac{d\mathbf{x}_\tau}{d \tau}=\begin{bmatrix}
        -10 & 10 & 0 \\
        28 & -1 &-\mathbf{x}[1]_\tau\\
        0 &\mathbf{x}[1]_\tau&-\frac{8}{3}
    \end{bmatrix} \mathbf{x}_\tau.
\end{equation}
Following the reasoning in \cite{KalmanNet,wellingHybridInference}, the transition matrix of the above equation has been assumed to remain constant over a short interval $\Delta\tau$; for consisteny with~\cite{KalmanNet}, $\Delta \tau=0.02$ was set. A discrete approximation of the continuous system described in~(\ref{eq:lorenzUpdate}) has been obtained via a truncated Taylor series expansion of the matrix exponential, yielding:
\begin{equation}\label{eq:lorenzNoisyUpdate}
\mathbf{x}_t = \mathbf{f}^{(j)}\left(\mathbf{x}_{t-1}\right) \mathbf{x}_{t-1} + \mathbf{w}_t^{(j)},
\end{equation}
where $\mathbf{f}^{(j)}(\cdot)$ was computed via a finite-order matrix exponential approximation dependent on the active system mode, and $\mathbf{w}^{(j)}_t$ denotes zero-mean Gaussian noise whose covariance matrix was set to $0.1 \mathbf{I}_3$ for all modes. 

The Lorenz Attractor system was assumed to switch between three discrete dynamic regimes, each with a different Taylor expansion order $J \in \{2,4,5\}$ and governed by the following Markov transition matrix:
\begin{equation}
    \boldsymbol{\Pi}= \begin{bmatrix}
        0.6 & 0.2 &0.2\\
        0.2 & 0.6 &0.2 \\
        0.2 & 0.2 &0.6
    \end{bmatrix}.
\end{equation}
Note that, for $J=2$, the system dynamics are approximately linear, while higher-order expansions lead to increasingly nonlinear evolution. In addition, we have considered that the observation structure varies across modes. In the first two modes, full-state noisy observations were provided. In the third mode, observations were first transformed into spherical coordinates before Gaussian noise inclusion. For all three modes, the level of the observation noise $\mathbf{v}^{(j)}_t$ was chosen to be the same varying from $-30$ to $10$~dB. 

\begin{figure*}[!t]
    \begin{subfigure}[b]{0.3\textwidth}
    \centering
        \includegraphics[width=\textwidth]{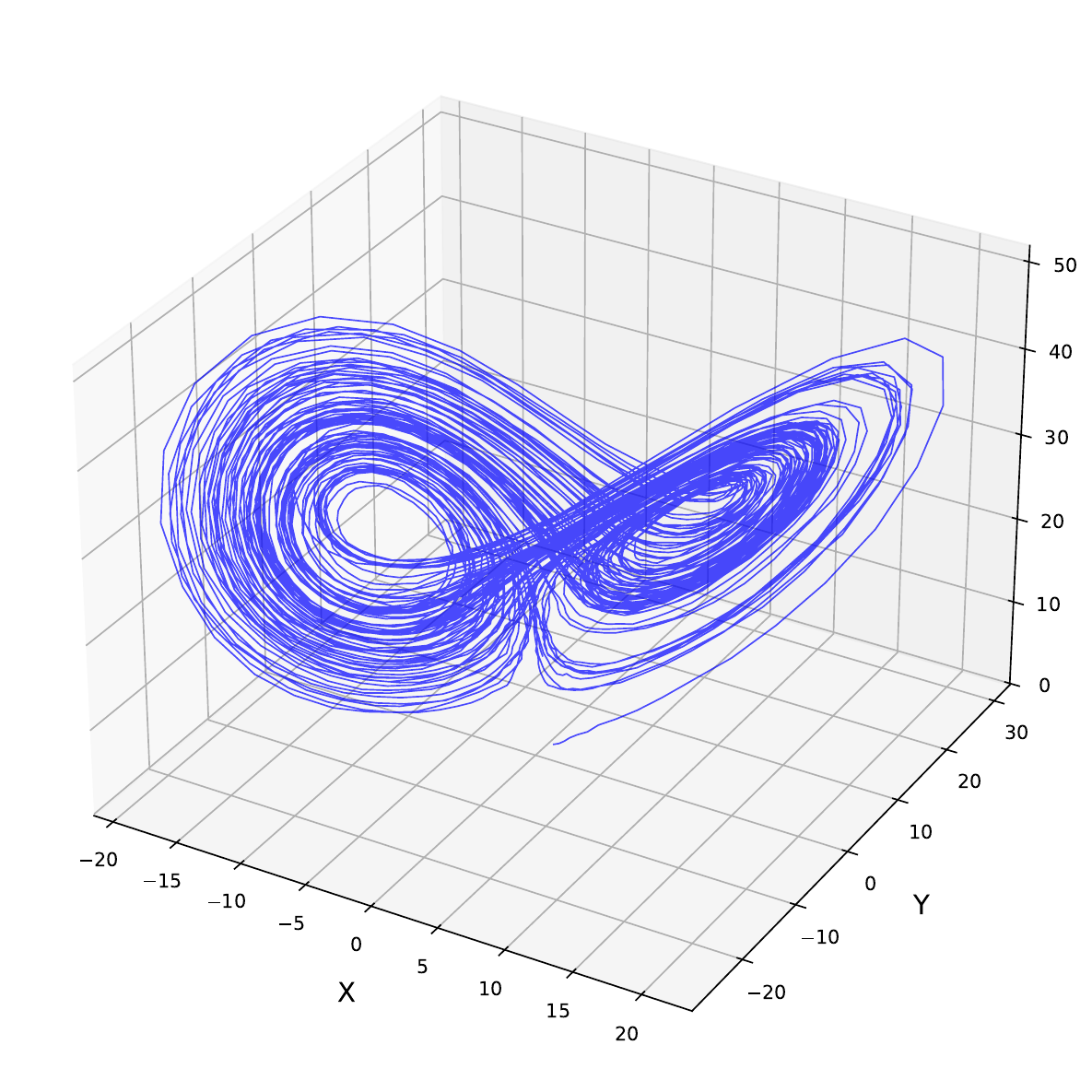}
        \caption{True state.}
        \label{fig:lorenz_state_evol}
    \end{subfigure}
    \begin{subfigure}[b]{0.3\textwidth}
        \centering
        \includegraphics[width=\textwidth]{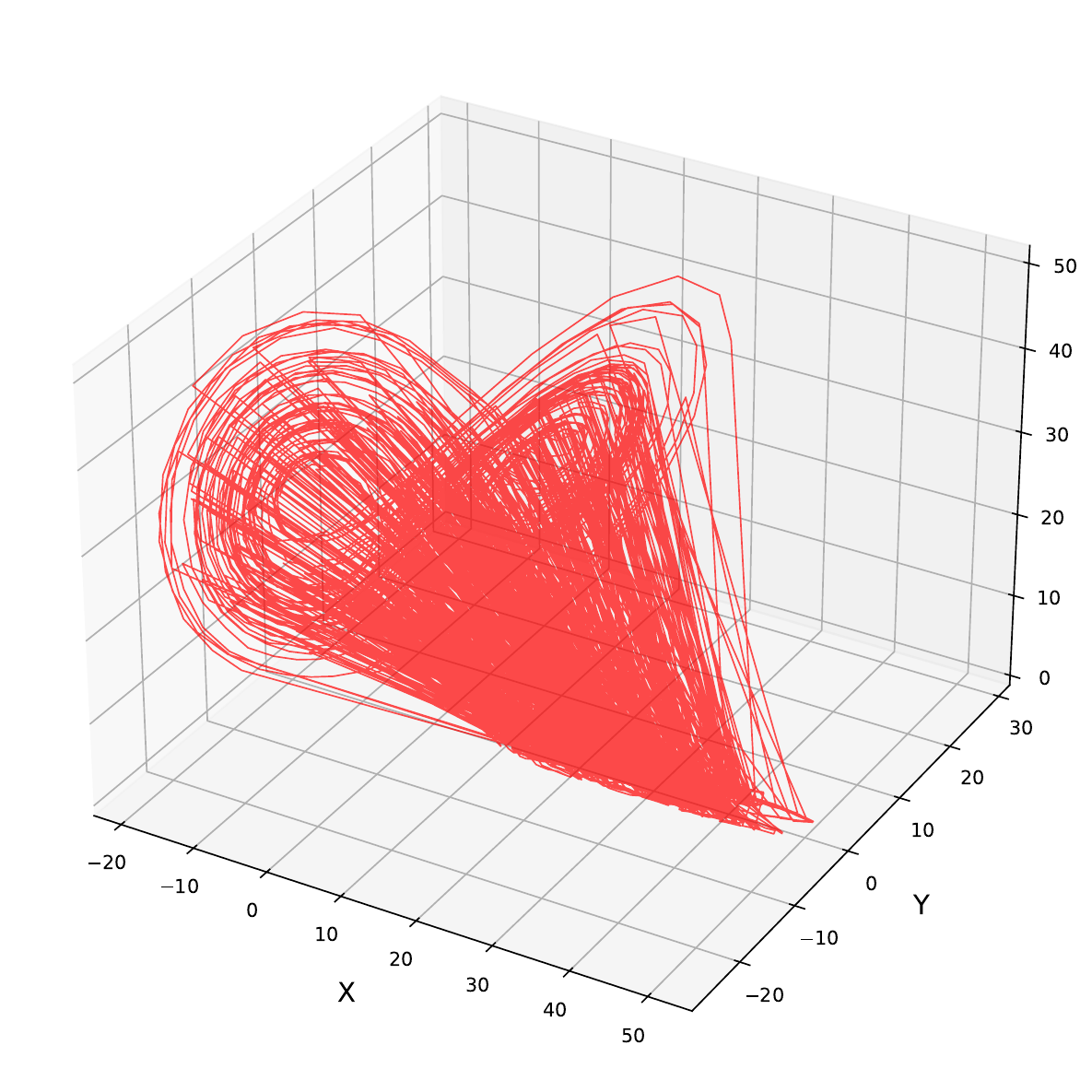} 
        \caption{Observation with $-20$~dB noise.}
        \label{fig:lorennz_obs_evol}
    \end{subfigure}
    \begin{subfigure}[b]{0.3\textwidth}
        \centering
        \includegraphics[width=1.25\textwidth]{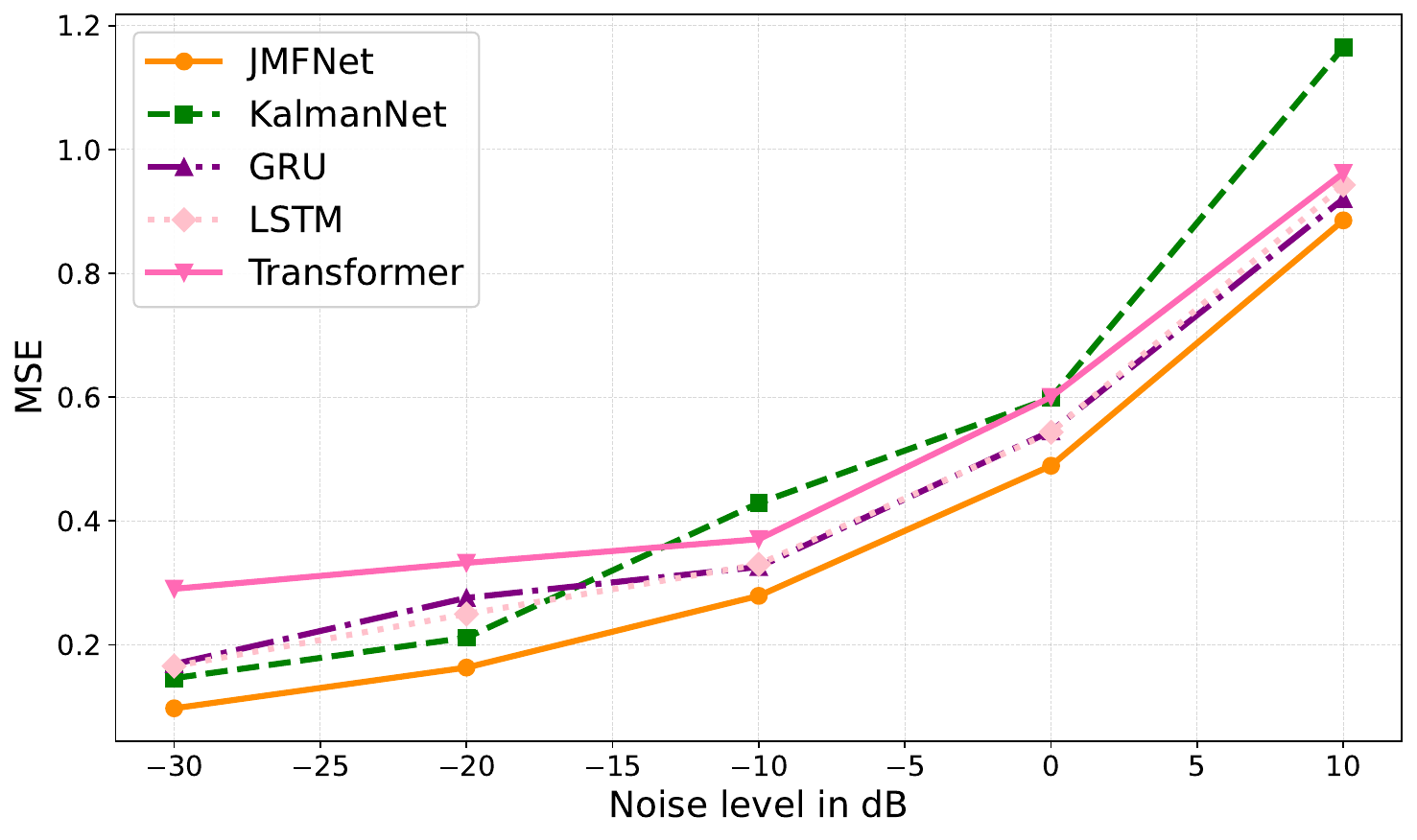} 
        \caption{Test set state estimation.}
        \label{fig:lorenz_resutlts}
    \end{subfigure}
    \caption{Example state and observation trajectories from the Lorenz attractor model in~(\ref{eq:lorenzNoisyUpdate}) with $M=3$ different Taylor expansion orders and different observation schemes for the modes, and the resulting test set performance for all considered state estimation schemes for different noise levels $\mathbf{v}^{(j)}_t$.}
    \label{fig:lorenz_plots}
\end{figure*}

Figures~\ref{fig:lorenz_state_evol} and~\ref{fig:lorennz_obs_evol} depict a state evolution example and the respective observations with noise power at $-20$~dB, whereas Fig.~\ref{fig:lorenz_resutlts} illustrates test set estimation performance for all implemented techniques; purely MB baselines achieved notably higher error probabilities hindering the clarity of the figure, hence, they have been excluded. It shown that JMFNet outperforms all baselines, with the KalmanNet performing well only in the low noise regime.

\subsection{Results for Traffic Monitoring}
The effectiveness of JMFNet against MF baselines has been also investigated over a real-world traffic forecasting task using the METR-LA dataset~\cite{metrLA}, including traffic speed measurements from $207$ sensors deployed across the Los Angeles County highway network. To this end, the system state at each timestep $t$ was represented by a $207$-dimensional vector of sensor readings, with the underlying dynamics exhibiting significant non-stationarity, varying considerably with the time of day due to congestion patterns, rush hours, and night-time traffic flows. To model those temporal variations, the daily timeline was discretized into the following four distinct operational modes: \textit{i}) morning rush (6–10~am), characterized by high-volume commuter traffic; \textit{ii}) midday (10~am to 3~pm), featuring a relative drop in congestion post rush hour; \textit{iii}) afternoon rush (3–7~pm), marked by increased traffic as commuters return from work; and \textit{iv}) night (7~pm to 6~am), defined by lower-volume, free-flowing traffic.

\begin{figure}
    \centering
    \includegraphics[width=\linewidth]{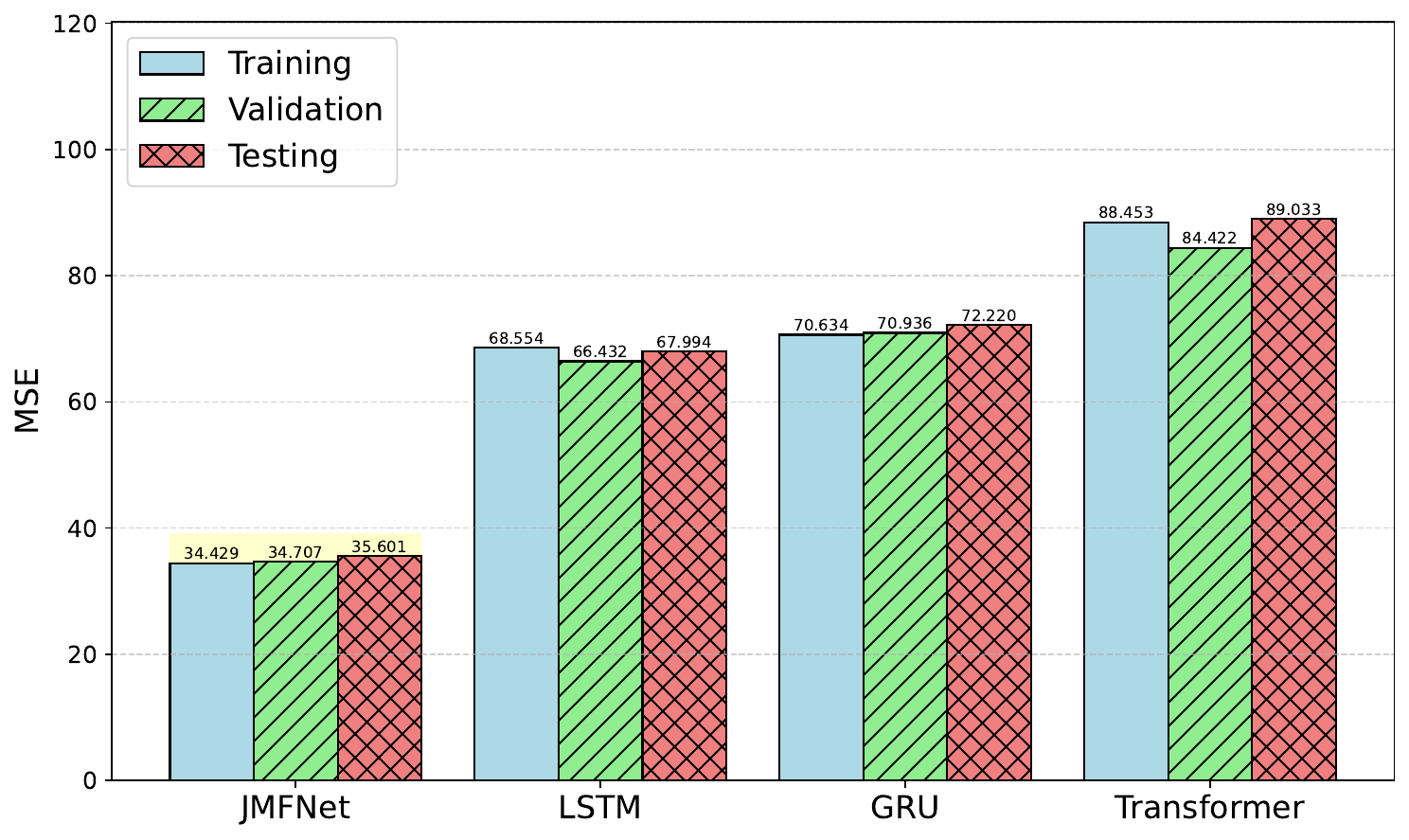}
    \caption{State estimation scores for the application of traffic modeling under partial observations considering a time horizon of $T=50$.}
    \label{fig:trafic_res}
\end{figure}

For each mode, a dedicated NN was trained to approximate the transition function $\mathbf{f}(\cdot)$ in~\eqref{eq:NonLDS_1}. Those NNs used $3$ ReLU activated hidden layers of $300$ units each, and were optimized using Adam with a learning rate of $0.001$. Noise-free partial observations were considered: all filters received as input readings from the first $100$ sensors, and were assigned to infer the entire state. The last $5000$ sequences were held out for testing. In addition, trajectories of length $T=50$ were generated by concatenating randomly sampled segments from the METR-LA dataset. Every $t_{\rm{tr}} \sim \rm{Uniform}[5,10]$ timesteps, one of four modes was uniformly selected and a corresponding sequence was sampled. A subsequence of length $t_{\rm tr}$ was extracted and appended to the growing trajectory. To maintain temporal consistency, each new segment was aligned 
to the end of the previous one via a simple translation offset. The state estimation scores for all schemes are included in~\ref{fig:trafic_res}, where it can be clearly seen that JMFNet outperforms all implemented MF baselines, achieving lower MSE across the training, validation, and test sets.

\section{Conclusion and Future Work}
In this paper, JMFNet, a novel MB DL approach incorporating domain knowledge with RNNs for real-time hidden state estimation in JMSs, was presented. In particular, the proposed filter 
combines partial model information with two moderately sized GRUs, which were trained using ALS. The presented performance evaluation over diverse applications showcased that our proposal outperforms traditional MB filters as well as purely MF RNNs and larger transformers. It was also shown that, on relatively simple tasks, the presented filter outperforms KalmanNet by a small but meaningful margin; on more challenging problems, such as chaotic systems with high observation noise, the performance gains become much more pronounced. In addition, it was demonstrated that JMFNet remains effective on sequences of long horizons, and that, ignoring mode switches (as in KalmanNet), can lead to diverging state estimates. This behavior renders the proposed filter particularly well-suited for high-stakes applications of JMss, such as finance, surveillance, and engineering control, where even modest improvements can translate into substantial practical benefits.

For future work, we aim to explore whether JMFNet can be combined with an additional CNN feature extractor \cite{latentKalmanNet}, or even multimodal learning techniques~\cite{multiModalSurvey}, to better handle high-dimensional observations with certain favorable structure. It is also worthwhile to investigate whether our online filtering framework, which leverages partially known dynamics, can benefit the inference capabilities of learned neural representations of SS models, such as those in \cite{deepKalmanFilters,foundationalMarkovJumpProc,wellingHybridInference}. Another a natural future direction is to extend JMFNet to other KF operations, such as smoothing~\cite{kalmanNetSmoothing}, missing observation recovery~\cite{neurIPsKF_timeSeries}, and control.

\bibliographystyle{ieeetr}
\bibliography{references}

\end{document}